\newif\ifisTR

\isTRtrue

\documentclass{article} 

\usepackage{fullpage}

\usepackage{microtype}
\usepackage{graphicx}
\usepackage{booktabs} 
\usepackage{placeins}
\usepackage{algorithm}
\usepackage{algorithmic}
\usepackage{subcaption}
\usepackage{tcolorbox}
\usepackage{nicefrac}  
\usepackage{enumitem}
\usepackage{multirow}
\usepackage{colortbl}
\usepackage{xspace}
\usepackage{wrapfig}
\usepackage{amsmath}
\usepackage{amssymb}
\usepackage{mathtools}
\usepackage[export]{adjustbox}
\usepackage{amsthm}
\usepackage{xcolor}

\definecolor{CColor}{rgb}{0.01,0.31,0.59}
\definecolor{GGray}{rgb}{0.80,0.90,1}
\definecolor{Shady}{rgb}{0.9,0.9,0.9}
\definecolor{kaistblue}{RGB}{20,135,200}
\definecolor{kaistdarkblue}{RGB}{0,65,145}
\definecolor{urbanablue}{RGB}{19,41,75}
\definecolor{urbanaorange}{RGB}{232,74,39}
\definecolor{drp}{rgb}{0.53,0.15,0.34}
\usepackage[plainpages=false,pdfpagelabels,colorlinks=true,linkcolor=CColor,citecolor=CColor,urlcolor=CColor]{hyperref}

\usepackage{tikz}

\usepackage[capitalize,noabbrev]{cleveref}

\newcommand{\ours}{LayerNorm Scaling\xspace}

\usepackage{overpic}

\theoremstyle{plain}
\newtheorem{theorem}{Theorem}[section]

\newtheorem{lemma}[theorem]{Lemma}

\theoremstyle{definition}

\newtheorem{assumption}[theorem]{Assumption}
\theoremstyle{remark}

\definecolor{mygray}{gray}{0.85}
\definecolor{LightBlue}{cmyk}{0.06, 0.03, 0.01, 0.0}

\usepackage[textsize=tiny]{todonotes}

\usepackage[round,semicolon,sort]{natbib}

\renewcommand{\cite}[1]{\citep{#1}}
\title{The Curse of Depth in Large Language Models}
\date{}

\author{
 Wenfang Sun$^*$\textsuperscript{1},
 Xinyuan Song$^*$\textsuperscript{2},
 Pengxiang Li$^*$\textsuperscript{3},
 Lu Yin\textsuperscript{4},
 Yefeng Zheng\textsuperscript{1},
 Shiwei Liu\textsuperscript{†5}
\\
 \textsuperscript{1}Westlake University, China \\
 \textsuperscript{2}Emory University, USA\\
 \textsuperscript{3}Dalian University of Technology, China \\
 \textsuperscript{4}University of Surrey, UK \\
  \textsuperscript{5}University of Oxford, UK \\
}

\begin{document}
\maketitle

\def\thefootnote{*}\footnotetext{Equal contribution. Accepted at NeurIPS 2025.}
\def\thefootnote{†}\footnotetext{Correspondence to: Shiwei Liu, shiwei.liu@maths.ox.ac.uk.}

\renewcommand{\thefootnote}{\arabic{footnote}}

\vspace{-3em}
\begin{abstract}

In this paper, we introduce \textit{the Curse of Depth}, a concept that highlights, explains, and addresses the recent observation in modern Large Language Models (LLMs) where nearly half of the layers are less effective than expected. We first confirm the wide existence of this phenomenon across the most popular families of LLMs such as Llama, Mistral, DeepSeek, and Qwen. Our analysis, theoretically and empirically, identifies that the underlying reason for the ineffectiveness of deep layers in LLMs is the widespread usage of Pre-Layer Normalization (Pre-LN). While Pre-LN stabilizes the training of Transformer LLMs, its output variance exponentially grows with the model depth, which undesirably causes the derivative of the deep Transformer blocks to be an identity matrix, and therefore barely contributes to the training. To resolve this training pitfall, we propose \textbf{LayerNorm Scaling (LNS)}, which scales the variance of output of the layer normalization inversely by the square root of its depth.\footnote{We found that combining LNS with Scaled Initialization \citep{groeneveld2024olmo,radford2019language,shoeybi2019megatron} diminishes the effectiveness of LNS. Therefore, we recommend removing the latter when applying LNS.} This simple modification mitigates the output variance explosion of deeper Transformer layers, improving their contribution. Across a wide range of model sizes (130M to 7B), our experiments show that LNS consistently outperforms previous normalization and scaling techniques in enhancing LLM pre-training performance. Moreover, this improvement seamlessly carries over to supervised fine-tuning. All these gains can be attributed to the fact that LayerNorm Scaling enables deeper layers to contribute more effectively during training. Our code is available at \href{https://github.com/lmsdss/LayerNorm-Scaling}{LayerNorm-Scaling}. 

\end{abstract}

\vspace{-0.5em}
\begin{figure*}[h]
  \centering
  \resizebox{0.9\textwidth}{!}{%
  \begin{minipage}[c]{0.49\linewidth}
    \centering
    \includegraphics[width=\linewidth]{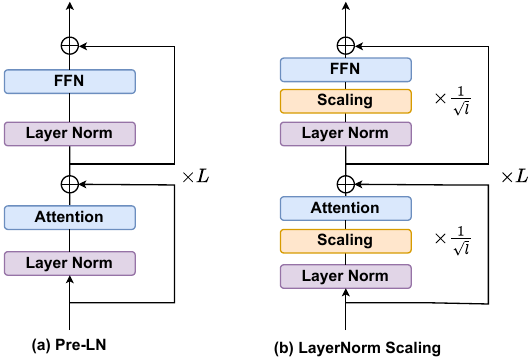}
  \end{minipage}\hfill
  \begin{minipage}[c]{0.49\linewidth}
    \centering
    \includegraphics[width=\linewidth]{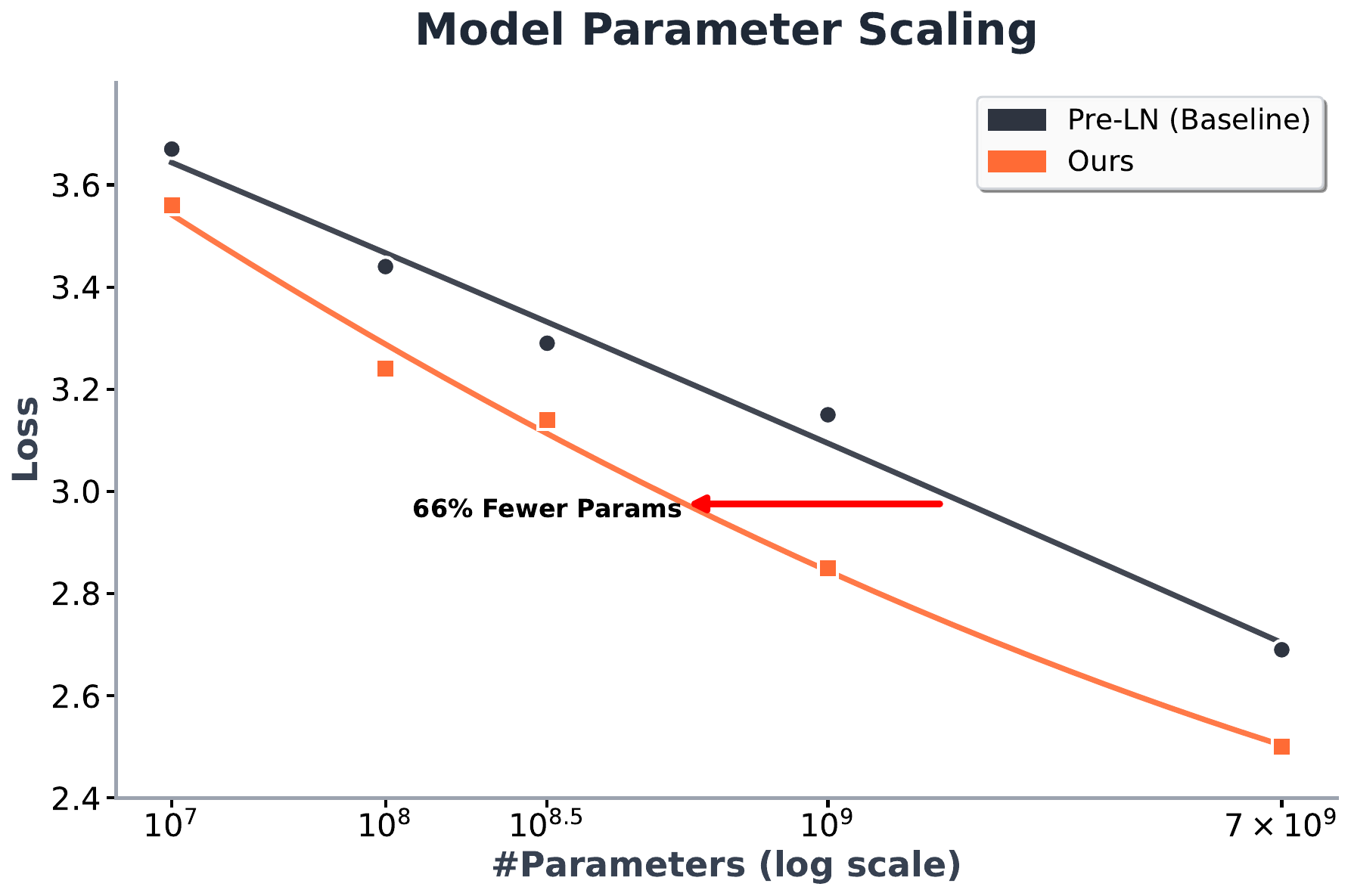}
  \end{minipage}}
 \caption{
  \textbf{Left:} Schematic diagrams of (a) Pre-LN and (b) LayerNorm Scaling. LayerNorm Scaling applies a scaling factor inversely proportional to the square root of the layer index $\ell$, preventing excessive variance growth. \textbf{Right}:  Language modeling loss of scaling up parameter count up to 7B. All models are trained for 20B tokens using OLMo \citep{groeneveld2024olmo}.}
  \label{fig:illus_scaling}
\end{figure*}

\clearpage
\tableofcontents

\clearpage
\section{Introduction}

Recent studies reveal that the deeper layers (Transformer blocks) in modern LLMs tend to be less effective than the earlier ones \citep{yin2023outlier,gromov2024unreasonable,men2024shortgpt,li2024mix}.
On the one hand, this interesting observation provides an effective indicator for LLM compression. For instance, we can compress deeper layers significantly more \citep{yin2023outlier,lu2024alphapruning,dumitru2024layer} to achieve high compression ratios. Even more aggressively, entire deep layers can be pruned completely without compromising performance \citep{muralidharan2024compact,siddiqui2024deeper}.


On the other hand, having many layers ineffective is undesirable as modern LLMs are extremely resource-intensive to train, often requiring thousands of GPUs trained for multiple months, let alone the labor used for data curation and administration \cite{achiam2023gpt,touvron2023llama}. Ideally, we want all layers in a model to be well-trained, with sufficient diversity
in features from layer to layer, to maximize the utility of resources \citep{li2024mix}. The existence of ill-trained layers suggests that there must be something off with current LLM paradigms. Addressing such limitations is a pressing need for the community to avoid the waste of valuable resources, as new versions of LLMs are usually trained with their previous computing paradigm which results in ineffective layers. 

To seek the immediate attention of the community, we re-introduce the concept of \textit{the Curse of Depth (CoD)} to systematically present the phenomenon of ineffective deep layers in various LLM families, to identify the underlying reason behind it, and to rectify it by proposing LayerNorm Scaling. We first state \textit{the Curse of Depth} below. 

\textbf{The Curse of Depth.} \textit{The Curse of Depth} refers to the observed phenomenon where deeper layers in modern LLMs contribute significantly less (\textit{but not nothing}) to learning and representation compared to earlier layers. These deeper layers often exhibit remarkable robustness to pruning and perturbations, implying they fail to perform meaningful transformations. This behavior prevents these layers from effectively contributing to training and representation learning, resulting in resource inefficiency.

\begin{figure*}[!t]
\centering
\includegraphics[width=1\linewidth]{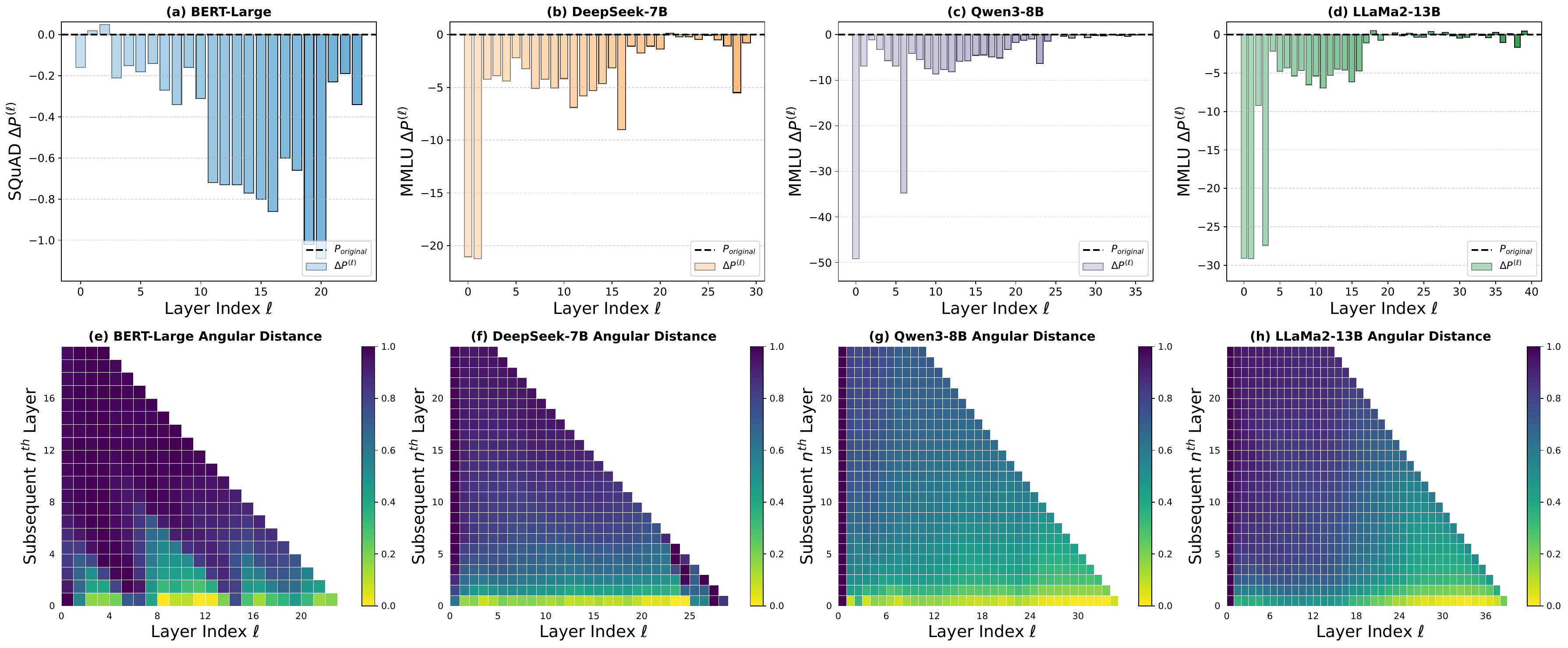}
\caption{Results of open-weight large-scale LLMs. \textbf{Top:} Performance drop after removing a single layer without fine-tuning.  
\textbf{Bottom:} Angular distance from the initial layer $\ell$ (x-axis) and its subsequent  $n^{\text{th}}$ layer (y-axis).  
The results demonstrate that in Pre-LN LLMs, deeper layers produce highly similar representations to their adjacent layers, and their removal results in minimal performance degradation. In contrast, Post-LN models show the opposite trend: deep layers contribute more substantially to model performance.}
\label{fig:performance_drop}
\end{figure*}

\textbf{Empirical Evidence of CoD.} To demonstrate that CoD is a common phenomenon across prominent LLM families, we perform layer pruning experiments on Qwen3, LLaMA2, and DeepSeek. Specifically, we prune one layer at a time, without any fine-tuning, and directly evaluate the resulting pruned models on the MMLU benchmark~\citep{hendrycks2020measuring}, as shown in Figure~\ref{fig:performance_drop}. 
\textbf{Key findings:} (1) Most models, including the latest Qwen3, exhibit surprising resilience to the removal of deeper layers; (2) The number of layers that can be removed without causing significant performance drop increases with model size; (3) Representations in deeper layers are significantly more similar to each other than those in earlier layers.

\textbf{Identifying the Root Cause of CoD.} We theoretically and empirically identify the root cause of CoD as the use of Pre-Layer Normalization (Pre-LN) \citep{baevski2018adaptive,dai2019transformer}, which normalizes layer inputs before applying the main computations, such as attention or feedforward operations, rather than after. Specifically, while stabilizing training, we observe that the output variance of Pre-LN accumulates significantly with layer depth as shown in Figure \ref{fig:combined_variance}, causing the derivatives of deep Pre-LN layers to approach an identity matrix. This behavior prevents these layers from introducing meaningful transformations, leading to diminished representation learning.  

\looseness=-1 \textbf{Mitigating CoD through LayerNorm Scaling.} We propose LayerNorm Scaling (LNS), which scales the output of Layer Normalization by the square root of the depth $\frac{1}{\sqrt{l}}$. LayerNorm Scaling effectively scales down the output variance across layers of Pre-LN. LNS consistently delivers better pre-training performance than existing normalization and scaling techniques across various model sizes from 130M to 7B. Unlike previous LayerNorm variants \citep{li2024mix, liu2020understanding}, LayerNorm Scaling is simple to implement, requires no hyperparameter tuning, and introduces no additional parameters during training. Furthermore, we show that the model pre-trained with LayerNorm Scaling achieves better performance on downstream tasks in self-supervised fine-tuning, all thanks to the more diverse feature representations learned in deep layers. 

\begin{figure}[h]
    \centering
    \hspace{-1em}
    \includegraphics[width=0.65\linewidth]{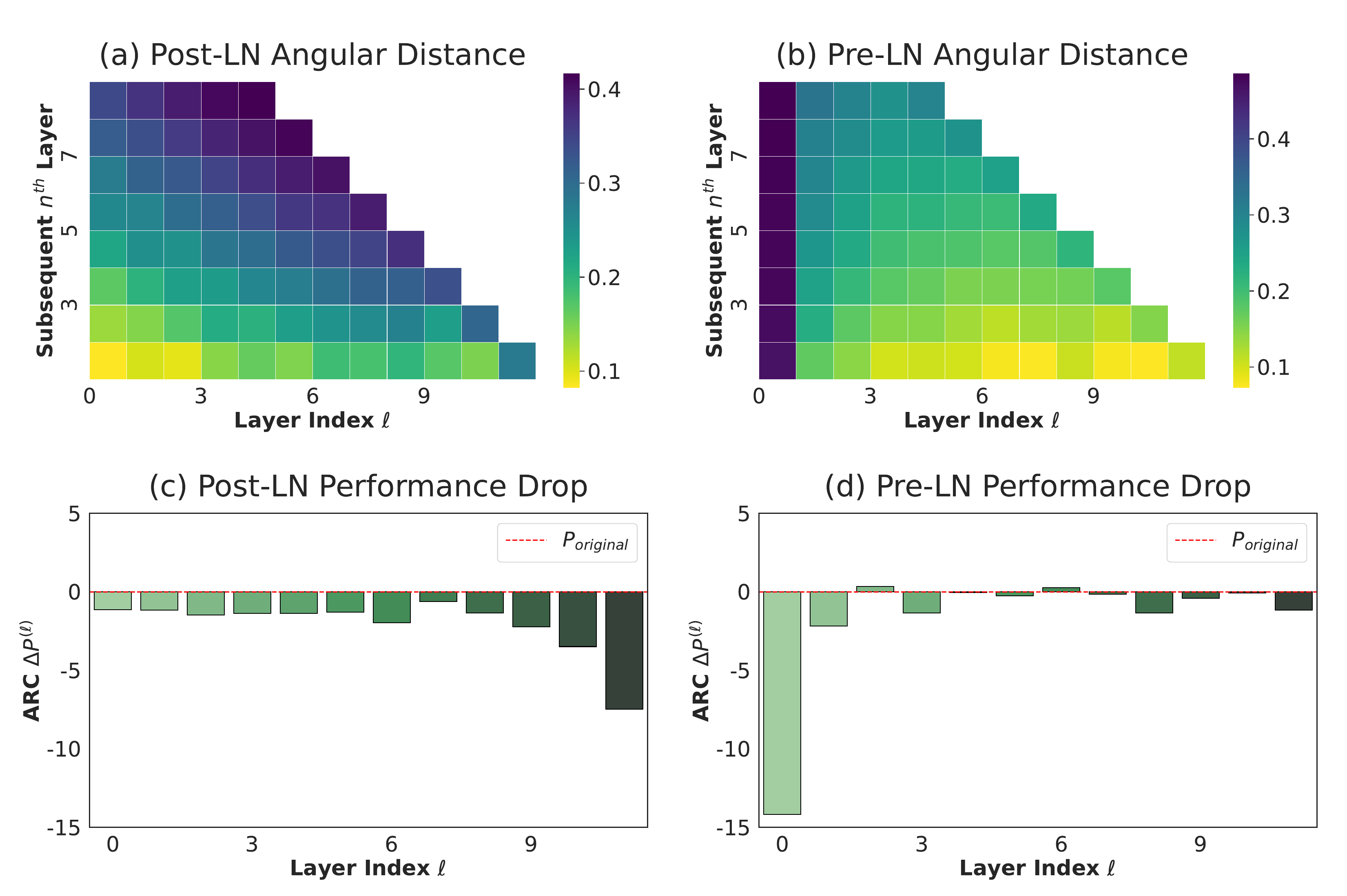}
    \caption{Results of in-house small-scale LLaMa-130M.  \textbf{Angular Distance (a, b)}: Each column represents the angular distance from the initial layer $\ell$ (x-axis) and its subsequent $n^{th}$ layer (y-axis). The distance is scaled to the range [0, 1], where yellow indicates smaller distances and purple indicates larger distances. \textbf{Performance Drop (c, d)}: ARC-e performance drop of removing each single layer from LLaMa-130M.}
    \label{fig:eval_in_house}
\end{figure}

\section{Empirical Evidence of the Curse of Depth} 

To empirically analyze the impact of layer normalization on the \textit{Curse of Depth} in LLMs, we conduct a series of evaluations inspired by \citet{li2024mix}, to compare Pre-LN and Post-LN models.


\textbf{Methodology:}  
We evaluate Pre-LN and Post-LN models by assessing the impact of layer pruning at different depths. Our hypothesis is that Pre-LN models exhibit diminishing effectiveness in deeper layers, whereas Post-LN models have less effective early layers.

\subsection{Open-weight Large-scale LLMs}

\textbf{Models:}  
To verify this, we empirically quantify the contribution of individual layers to overall model performance across a diverse set of LLMs, including Qwen3 \citep{qwen3_blog}, LLaMA2~\citep{touvron2023llama},  DeepSeek \citep{bi2024deepseek}, and BERT-Large~\citep{devlin2018bert}.
These models were chosen to ensure architectural and application diversity. BERT-Large represents a Post-LN model, whereas the rest are Pre-LN-based. This selection enables a comprehensive evaluation of the effects of layer normalization across varying architectures and model scales.

\textbf{Evaluation Metric:}  
To empirically assess the impact of deeper layers in LLMs, we adopt two metrics, \textit{Performance Drop}  and \textit{Angular Distance}, inspired by \citet{gromov2024unreasonable,li2024mix}. 

\textit{Performance Drop} \( \Delta P^{(\ell)} \) quantifies the importance of each layer by measuring the performance change after its removal. A smaller \( \Delta P^{(\ell)} \) indicates that the pruned layer contributes less to the model’s overall performance. For BERT-Large, we evaluate using the SQuAD v1.1 dataset~\citep{rajpurkar2016squad}, while for other models, we use MMLU~\citep{hendrycks2020measuring}, a standard benchmark for multi-task language understanding.

\textit{Angular Distance} \( d(x^{\ell}, x^{\ell+n}) \) quantifies the directional change between the input representations at layer \( \ell \) and layer \( \ell + n \) on a neutral pre-training dataset. Formally, given a token \( T \), let \( x_T^{\ell} \) and \( x_T^{\ell+n} \) denote its input to layers \( \ell \) and \( \ell + n \), respectively. The angular distance is defined as:
\begin{equation}\label{eq:arccos-sim}
d(x^{\ell}, x^{\ell+n}) = \frac{1}{\pi} \arccos\left(\frac{x_T^{\ell} \cdot x_T^{\ell+n}}{\|x_T^{\ell}\|_2 \|x_T^{\ell+n}\|_2} \right),
\end{equation}
where \( \|\cdot\|_2 \) denotes the \( L^2 \)-norm. To reduce variance, we report the average distance over 256K tokens sampled from the C4 dataset. Smaller values of \( d(x^{\ell}, x^{\ell+n}) \) indicate higher similarity between the two representations, suggesting limited transformation. Such layers can be considered redundant, as their removal minimally impacts the model's internal representations. Ideally, each layer should introduce meaningful representational shifts to fully leverage the model’s capacity~\citep{yang2023tensor,gromov2024unreasonable}.

\textbf{Experimental Results:}  
(1) Pruning deep layers in Pre-LN LLMs leads to negligible, and sometimes even positive, changes in performance, as shown in Figure~\ref{fig:performance_drop}-Top.  
Specifically, Figure~\ref{fig:performance_drop} (b)--(d) reveals that a wide range of deeper layers—particularly beyond the 18th—can be pruned with minimal impact on performance. This indicates that deep layers in Pre-LN architectures contribute little to the model’s overall effectiveness.  
In contrast, Figure~\ref{fig:performance_drop} (a) shows that pruning deep layers in BERT-Large (a Post-LN model) leads to a substantial drop in accuracy, while pruning early layers has a relatively minor effect.
(2) Pre-LN models exhibit decreasing angular distance in deeper layers, indicating highly similar representations, as shown in Figure~\ref{fig:performance_drop}-Bottom. For instance, the angular distance in DeepSeek-7B falls below 0.2 after the 18th layer. Qwen3-8B demonstrates a higher similarity, with nearly half of its layers exhibiting distances below 0.2 from their preceding layers. In LLaMA2-13B, the angular distance approaches zero across the final one-third of the network. These similar representations align with the pruning results in Figures~\ref{fig:performance_drop} (b)--(d), where pruning later layers has little effect, while pruning early layers significantly degrades performance.

\begin{figure*}[t]
\centering
\includegraphics[width=0.9\linewidth]{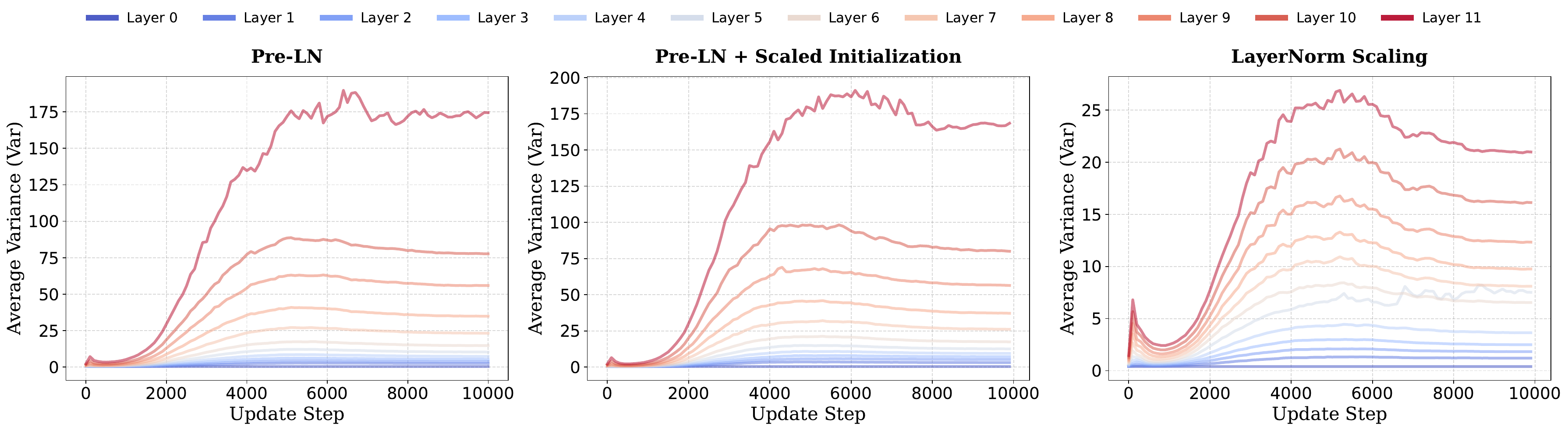}
\caption{
\textbf{Layerwise output variance.} This figure compares the output variance across various layers for different setups: (1) Pre-LN; (2) Pre-LN with Scaled Initialization \citep{shoeybi2019megatron,radford2019language}; and (3) LayerNorm Scaling. The experiments are conducted on the LLaM-130M model trained for 10,000 steps. The proposed \ours{} effectively controls the variance across layers.
}
\label{fig:combined_variance}
\end{figure*}

\subsection{In-house Small-scale LLaMa-130M}
\label{app:control_exp}

To eliminate the influence of other confounding variables, we train two LLaMA-130M models from scratch that differ only in their Layer Normalization, thereby clearly distinguishing Post-LN from Pre-LN, following \citet{li2024mix}. The results are illustrated in Figure \ref{fig:eval_in_house}. 

In Post-LN models, early layers exhibit high similarity (low angular distance, Figure \ref{fig:eval_in_house}-a) and their removal causes minimal performance loss (Figure~\ref{fig:eval_in_house}-c), while deeper layers become more distinct and critical. Conversely, Pre-LN LLaMa-130M demonstrates a gradual decrease in angular distance with depth, resulting in highly similar deep layers (Figure~\ref{fig:eval_in_house}-b). Removing most layers after the first in Pre-LN causes negligible performance loss (Figure~\ref{fig:eval_in_house}-d), indicating their limited contribution. These consistent findings, observed in both open-weight and in-house LLMs, lead to the conclusion that the widespread use of Pre-LN is the root cause of the ineffectiveness of deep layers in LLMs.

\section{Analysis of the Curse of Depth}
\textbf{Preliminaries.} This paper primarily focuses on Pre-LN Transformer \citep{baevski2018adaptive,dai2019transformer}. Let $x_\ell \in \mathbb{R}^d$ be the input vector at the $\ell$-th layer of Transformer, where $d$ denotes the feature dimension of each layer. For simplicity, we assume all layers to have the same dimension $d$. The layer output $y$ is calculated as follows:
\begin{equation}
    y = x_{\ell+1} = x^\prime_\ell + \mathrm{FFN}(\mathrm{LN}(x^\prime_\ell)),  
    \label{x_result}
\end{equation}
\begin{equation}
    x^\prime_\ell = x_\ell + \mathrm{Attn}(\mathrm{LN}(x_\ell)),
    \label{xprime_result}
\end{equation}
where LN denotes the layer normalization function. In addition, the feed-forward network (FFN) and the multi-head self-attention (Attn) sub-layers are defined as follows:
\begin{equation}
\begin{aligned}
    \mathrm{FFN}(x) &= W_2 \mathcal{F}(W_1 x), \\
    \mathrm{Attn}(x) &= W_O (\mathrm{concat}(\mathrm{head}_1(x), \dots, \mathrm{head}_h(x))),  \\
    \mathrm{head}_i(x) &= \mathrm{softmax} \left( \frac{(W_{Qi} x)^\top (W_{Ki} X)}{\sqrt{d_\mathrm{head}}} \right) (W_{Vi} X)^\top,
\end{aligned}
\end{equation}
where $\mathcal{F}$ is an activation function, $\mathrm{concat}$ concatenates input vectors, $\mathrm{softmax}$ applies the softmax function, and $W_1 \in \mathbb{R}^{d_\mathrm{ffn} \times d}$, $W_2 \in \mathbb{R}^{d \times d_\mathrm{ffn}}$, $W_{Qi} \in \mathbb{R}^{d_\mathrm{head} \times d}$, $W_{Ki} \in \mathbb{R}^{d_\mathrm{head} \times d}$, $W_{Vi} \in \mathbb{R}^{d_\mathrm{head} \times d}$, and $W_O \in \mathbb{R}^{d \times d}$ are parameter matrices, and $d_\mathrm{FFN}$ and $d_\mathrm{head}$ are the internal dimensions of FFN and multi-head self-attention sub-layers, respectively. $X \in \mathbb{R}^{d \times s}$, where $s$ is the input sequence length.

The derivatives of Pre-Ln Transformers are:
\begin{equation}
\frac{\partial \text{Pre-LN}(x)}{\partial x} = I + \frac{\partial f (\text{LN}(x))}{\partial \text{LN}(x)} \frac{\partial \text{LN}(x)}{\partial x},
\end{equation}
where $f$ here represents either the multi-head attention function or the FFN function. If the term $\frac{\partial f (\text{LN}(x))}{\partial \text{LN}(x)} \frac{\partial \text{LN}(x)}{\partial x}$ becomes too small, the Pre-LN layer $\frac{\partial \text{Pre-LN}(x)}{\partial x}$ behaves like  an identity map. Our main objective is to prevent identity map behavior for very deep Transformer networks. The first step in this process is to compute the variance $\sigma^2_{x_\ell}$ of vector $x_{\ell}$.

\subsection{Pre-LN Transformers}

\begin{assumption}
Let $x_\ell$ and $x^\prime_\ell$ denote the input and intermediate vectors of the $\ell$-th layer. Moreover, let $W_\ell$ denote the model parameter matrix at the $\ell$-th layer. We assume that, for all layers, $x_\ell$, $x^\prime_\ell$, and $W_\ell$ follow normal and independent distributions with mean $\mu = 0$.
\label{total_assumption}
\end{assumption}

\begin{lemma}
\label{attention_variance}
Let $\sigma^2_{x^\prime_\ell}$ and $\sigma^2_{x_\ell}$ denote the variances of $x^\prime_\ell$ and $x_\ell$, respectively. These two variances exhibit the same overall growth trend, which is:
\begin{equation}
\sigma^2_{x_{\ell}} = \sigma_{x_1}^2 \Theta\Bigl(\prod_{k=1}^{\ell-1} \left( 1 + \frac{1}{\sigma_{x_k}} \right) \Bigr),
\label{sigma_l}
\end{equation}
where the growth of $\sigma^2_{x_\ell}$ is sub-exponential, as shown by the following bounds:
\begin{equation}
\Theta(L) \leq \sigma^2_{x_L} \leq \Theta(\exp(L)).
\end{equation}
\label{convergence_of_variance}
\end{lemma}

Here, the notation $\Theta$ means: if $f(x) \in \Theta\bigl(g(x)\bigr)$, then there exist constants $C_1, C_2$ such that $C_1|g(x)| \le |f(x)| \le C_2|g(x)|$ as $x \to \infty$. The lower bound $ \Theta(L) \leq \sigma^2_{x_\ell} $ indicates that $\sigma^2_{x_\ell}$ grows at least linearly, while the upper bound $\sigma^2_{x_\ell} \leq \Theta(\exp(L))$ implies that its growth does not exceed an exponential function of $L$.

\begin{figure}[t]
    \centering
    \includegraphics[width=\linewidth]{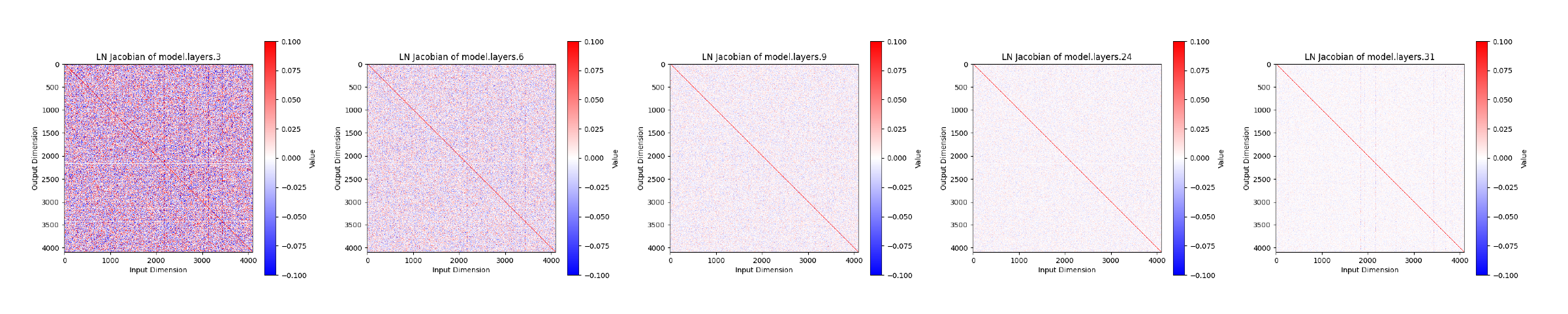}
    \caption{
    Visualization of the Jacobian matrices of pre-LN residual blocks across different layers of a pre-trained
    LLaMA2-7B model.
    Each heatmap shows the token-averaged Jacobian at a specific layer.
    As depth increases, the Jacobians exhibit a pronounced diagonal dominance with vanishing
    off-diagonal entries, indicating that deep LayerNorm blocks increasingly approximate
    identity mappings.
    }
    \label{fig:llama2_jacobian}
\end{figure}

Based on Assumption \ref{total_assumption} and the work of \cite{takase2023spike}, we obtain the following:
\begin{theorem}
For a Pre-LN Transformer with $L$ layers, using Equations~\eqref{x_result} and \eqref{xprime_result}, the partial derivative $\frac{\partial y_L}{\partial x_1}$ can be written as:
\begin{equation}
\frac{\partial y_L}{\partial x_1} = \prod_{\ell=1}^{L-1} \left( \frac{\partial y_\ell}{\partial x^\prime_\ell} \cdot \frac{\partial x^\prime_\ell}{\partial x_\ell} \right).
\end{equation}

The Euclidean norm of $\frac{\partial y_L}{\partial x_1}$ is given by:
\begin{equation}
\left\| \frac{ \partial y_L}{\partial x_1} \right\|_2 \leq \prod_{l=1}^{L-1} \left( 1 + \frac{1}{\sigma_{x_\ell}} A + \frac{1}{\sigma_{x_\ell}^2} B \right),
\label{traditional_form}
\end{equation}
where $A$ and $B$ are constants for the Transformer network. Then the upper bound for this norm is given as follows: when  $\sigma^2_{x_\ell}$ grows exponentially, (i.e., at its upper bound), we have:
\begin{equation}
    \sigma^2_{x_\ell} \sim \exp(\ell), \quad \left\| \frac{\partial y_L}{\partial x_1} \right\|_2 \leq M,
\end{equation}
where the gradient norm converges to a constant $M$. Conversely, when $\sigma^2_{x_{\ell}}$ grows linearly (i.e., at its lower bound), we have
\begin{equation}
    \sigma^2_{x_\ell} \sim \ell, \quad \left\| \frac{\partial y_L}{\partial x_1} \right\|_2 \leq \Theta(L),
\end{equation}
which means that the gradient norm grows linearly in $L$.
\label{main_result}
\end{theorem}
The detailed description of $A$ and $B$, as well as the complete proof, are provided in Appendix \ref{appendix:proof2}.
From Theorem \ref{main_result}, we observe that when the variance grows exponentially, as the number of layers $L \to \infty$, the norm $\left\| \frac{\partial y_L}{\partial x_1} \right\|_2$ is bounded above by a fixed constant $M$. This result implies that even an infinitely deep Transformer remains stable, and by the Weierstrass Theorem, the network is guaranteed to converge.
Consequently, this implies that for very large $L$, deeper layers behave nearly as an \textbf{identity map} from $x_\ell$ to $y_\ell$, thereby limiting the model’s expressivity and hindering its ability to learn meaningful transformations. This phenomenon is empirically illustrated in Figure~\ref{fig:llama2_jacobian}, 
where we visualize the Jacobian of the pre-LN residual blocks across depth in a pre-trained LLaMA2-7B model, 
revealing a clear collapse toward identity mappings in deeper layers. This outcome is undesirable, therefore, we would instead prefer the variance to increase more gradually—e.g., linearly—so that $\left\| \frac{\partial y_L}{\partial x_1} \right\|_2$ exhibits linear growth. This observation highlights the necessity of appropriate variance control mechanisms, such as scaling strategies, to prevent excessive identity mappings and enhance network depth utilization.

\section{LayerNorm Scaling (LNS)} 

To mitigate the abovementioned issue, we propose \ours, a simple yet effective normalization strategy. The core idea of \ours is to control the exponential growth of output variance in Pre-LN by scaling the normalized outputs according to layer depth. Specifically, we apply a scaling factor inversely proportional to the square root of the layer index to scale down the output of LN layers, enhancing the contribution of deeper Transformer layers during training. LayerNorm Scaling is illustrated in Figure \ref{fig:illus_scaling}. 


Formally, for a Transformer model with $L$ layers, the output of Layer Normalization in each layer $\ell$ is scaled by a factor of $\frac{1}{\sqrt{\ell}}$. Let $\mathbf{h}^{(\ell)}$ denote the input to Layer Normalization at layer $\ell$. The modified output is computed as:
\begin{equation}
\mathbf{h}^{(\ell)} = \text{LayerNorm}(\mathbf{h}^{(\ell)}) \times \frac{1}{\sqrt{\ell}},
\end{equation}
where $\ell \in \{1, 2, \dots, L\}$. This scaling prevents excessive variance growth with depth, addressing a key limitation of Pre-LN. Unlike Mix-LN, which stabilizes gradients in deeper layers but suffers from training instability caused by Post-LN \citep{nguyen2019transformers,wang2024deepnet}, \ours preserves the stability advantages of Pre-LN while enhancing the contribution of deeper layers to representation learning. Applying \ours leads to a notable reduction of layerwise output variance as shown in Figure \ref{fig:combined_variance}, resulting in a lower training loss. Moreover, compared with previous LayerNorm variants \citep{li2024mix,liu2020understanding}, \ours is hyperparameter-free, easy to implement, and does not introduce additional learnable parameters, making it computationally efficient and readily applicable to existing Transformer architectures. 

\subsection{Theoretical Analysis of \ours}
\label{app:analysis_lNS}
\begin{lemma}
\label{scaling_attention_variance}
After applying our scaling method, the variances of $x^\prime_\ell$ and $x_\ell$, denoted as $\sigma^2_{x^\prime_\ell}$ and $\sigma^2_{x_\ell}$, respectively, exhibit the same growth trend, which is:
\begin{equation}
\begin{aligned}
\sigma^2_{x_{\ell}} =\sigma_{x_1}^2 \Theta\Big( \prod^{\ell-1}_{k=1}\Big(1 + \frac{1}{\sqrt{k} \sigma_{x_k}}\Big)\Big),
\label{sigma_l_scaling}
\end{aligned}
\end{equation}
with the following growth rate bounds:
\begin{equation}
\Theta(L) \leq \sigma^2_{x_L} \leq \Theta(L^{(2-\epsilon)}).
\end{equation}
where $\epsilon$ is a small number with $1/2 \leq \epsilon < 1$. 
\label{convergence_of_variance_scaling}
\end{lemma}

From Lemma \ref{convergence_of_variance_scaling}, we can conclude that our scaling method effectively slows the growth of the variance upper bound, reducing it from exponential to polynomial growth. Specifically, it limits the upper bound to a quadratic rate instead of an exponential one. Based on Theorem \ref{main_result}, after scaling, we obtain the following:

\begin{theorem}

For the scaled Pre-LN Transformers, the Euclidean norm of $\frac{\partial y_L}{\partial x_1}$ is given by:
\begin{equation}
\left\| \frac{ \partial y_L}{\partial x_1} \right\|_2 \leq \prod_{\ell=1}^{L-1} \left( 1 + \frac{1}{\ell\sigma_{x_\ell}} A + \frac{1}{\ell^2\sigma_{x_\ell}^2} B \right),
\label{traditional_form_scaling}
\end{equation}
where $A$ and $B$ are dependent on the scaled neural network parameters. Then the upper bound for the norm is given as follows: when $\sigma^2_{x_\ell}$ grows at $\ell^{(2-\epsilon)}$, (i.e., at its upper bound), we obtain:
\begin{equation}
    \sigma^2_{x_\ell} \sim \ell^{(2-\epsilon)}, \quad \left\| \frac{\partial y_L}{\partial x_1} \right\|_2 \leq \omega(1),
\end{equation}
where $\omega$ denotes that if $f(x) = \omega(g(x))$, then $\lim_{x \rightarrow \infty} \frac{f(x)}{g(x)} = \infty$. Meanwhile, when $\sigma^2_{x_{\ell}}$ grows linearly (i.e., at its lower bound), we obtain:
\begin{equation}
    \sigma^2_{x_\ell} \sim \ell, \quad \left\| \frac{\partial y_L}{\partial x_1} \right\|_2 \leq \Theta(L).
\end{equation}
\label{main_result_scaling}
\end{theorem}

\looseness=-1 The detailed descriptions of $A$ and $B$, and $\epsilon$, along with the full proof, are provided in Appendices \ref{appendix:proof3} and  \ref{appendix:proof4}.

By comparing Theorem \ref{main_result} (before scaling) with Theorem \ref{main_result_scaling} (after scaling), we observe a substantial reduction in the upper bound of variance. Specifically, it decreases from exponential growth $\Theta(\exp(L))$ to at most quadratic growth $\Theta(L^2)$. In fact, this growth is even slower than quadratic expansion, as it follows $\Theta(L^{(2-\epsilon)})$ for some small $\epsilon > 0$.

When we select a reasonable upper bound for this expansion, we find that \( \left\| \frac{\partial y_L}{\partial x_1} \right\|_2 \) no longer possesses a strict upper bound. That is, as the depth increases, \( \left\| \frac{\partial y_L}{\partial x_1} \right\|_2 \) continues to grow gradually.
Consequently, fewer layers act as identity mappings compared to the original Pre-LN where nearly all deep layers collapsed into identity transformations. Instead, the after-scaled network effectively utilizes more layers, even as the depth approaches infinity, leading to improved expressivity and trainability.

In addition to making the deeper layers more effective, our variance-scaling approach can also reduce sudden spikes in the loss landscape during training. Based on~\cite{takase2023spike}'s work, We formalize this in the following theorem Theorem~\ref{thm:gradient-bound-lossspike}, which gives a rigorous upper bound on the gradient norm with respect to the attention parameters.

\begin{theorem}
\label{thm:gradient-bound-lossspike}
For the Pre-Ln transformers with weight $W_1$ on its first layer's query projection. Then the $L$-layer backpropagated gradient norm with respect to $W_1$ satisfies the following upper bound:
\begin{equation}
\left\| \frac{ \partial y_L}{\partial W_1} \right\|_2 \leq \prod_{\ell=1}^{L-1} \left( 1 + \frac{1}{\ell\sigma_{x_\ell}} A^\prime + \frac{1}{\ell^2\sigma_{x_\ell}^2} B^\prime \right),
\label{traditional_form_scaling2}
\end{equation}
where $A^\prime$ and $B^\prime$ are dependent on the scaled neural network parameters defined in~\ref{proof5}.
\end{theorem}

From~\eqref{traditional_form_scaling2}, we can easily get that if we do not want so many loss spikes, we need to let the $\left\|\frac{\partial y_L}{\partial W_1}\right\|_2$ do not explode. Which in our assumption means that the variance of the deep layer should not be too small.
Based on the above result~\eqref{traditional_form_scaling}, the good variance growth rate is sub linearly growth. which is: 
\begin{equation}
\sigma^2_{x_\ell} \sim \ell,
\end{equation}
which is actually the \ours convergence rate. Therefore, the \ours method can provide a moderate scaling of the variance, both to make the deeper layers effective and to prevent the initial layers from exploding.

The proof of Theorem~\ref{thm:gradient-bound-lossspike} is in Section~\ref{proof5}. Then we can easily generalize to a more general situation for layer $l$. By carefully controlling the propagation of gradients through the attention blocks, we can observe that for every layer $\ell$, the $\left\| \frac{ \partial y_L}{\partial W_\ell} \right\|_2$ has the same upper bound as in Result~\ref{traditional_form_scaling2}. The proof is omitted here. As a result, \ours improves stability for every layer (especially the first layer) and avoids sharp gradient amplification, which would otherwise result in an unstable or inefficient optimization process.

\section{Experiments}%

\subsection{LLM Pre-training}

\label{sec:pre}
To evaluate the effectiveness of \ours, we follow the experimental setup of \citet{li2024mix}, using the identical model configurations and training conditions to compare LNS with widely used normalization techniques, including Post-LN~\citep{nguyen2019transformers}, DeepNorm~\citep{wang2024deepnet}, and Pre-LN~\citep{dai2019transformer}. In line with \citet{lialin2023relora} and \citet{zhao2024galore}, we conduct experiments using LLaMA-based architectures with model sizes of 130M, 250M, 350M, and 1B parameters.

\begin{table*}[!ht]
\centering
\caption{Perplexity (↓) comparison of various layer normalization methods.}

\label{tab:norm_comparison}
\resizebox{0.9\textwidth}{!}{%
\begin{tabular}{lcccc}
\toprule
& \textbf{LLaMA-130M} & \textbf{LLaMA-250M} & \textbf{LLaMA-350M} & \textbf{LLaMA-1B}\\ 
Training Tokens  & 2.2B & 3.9B & 6.0B & 8.9B \\
\midrule
Post-LN \citep{ba2016layer}  &  26.95 & 1409.79 & 1368.33 & 1390.75 \\  
DeepNorm \citep{wang2024deepnet} & 27.17 & 22.77 & 1362.59 & 1409.08 \\
Mix-LN \citep{li2024mix}  &  26.07 & 21.39 & 1363.21 & 1414.78\\
Pre-LN \citep{baevski2018adaptive} & 26.73 & 21.92 & 19.58 & 17.02 \\
\midrule
Pre-LN + LayerNorm Scaling  & \textbf{25.76} & \textbf{20.35} & \textbf{18.20} & \textbf{15.71}\\
\bottomrule
\end{tabular}}
\end{table*}

The architecture incorporates RMSNorm \citep{shazeer2020glu} and SwiGLU activations \citep{zhang2019root}, which are applied consistently across all model sizes and normalization methods. For optimization, we use the Adam optimizer \citep{kingma2014adam} and adopt size-specific learning rates: $1 \times 10^{-3}$ for models up to 350M parameters, and $5 \times 10^{-4}$ for the 1B parameter model. All models share the same architecture, hyperparameters, and training schedule, with the only difference being the choice of normalization method. Unlike Mix-LN \citep{li2024mix}, which introduces an additional hyperparameter $\alpha$ manually set to 0.25, \ours requires no extra hyperparameters, making it simpler to implement.
Table~\ref{tab:norm_comparison} shows that \ours consistently outperforms other normalization methods across different model sizes. While DeepNorm performs comparably to Pre-LN on smaller models, it struggles with larger architectures like LLaMA-1B, showing signs of instability and divergence in loss values. Similarly, Mix-LN outperforms Pre-LN in smaller models but faces convergence issues with LLaMA-350M, indicating its sensitivity to architecture design and hyperparameter tuning due to the introduction of Post-LN. Notably, Mix-LN was originally evaluated on LLaMA-1B with 50K steps \citep{li2024mix}, while our setting extends training to 100K steps, where Mix-LN fails to converge, highlighting its instability in large-scale settings caused by the usage of Post-LN.

\looseness=-1 In contrast, \ours solves the \textit{Curse of Depth} without compromising the training stability. \ours achieves the lowest perplexity across all tested model sizes, showing stable performance improvements over existing methods. For instance, on LLaMA-130M and LLaMA-1B, \ours reduces perplexity by 0.97 and 1.31, respectively, compared to Pre-LN. Notably, \ours maintains stable training dynamics for LLaMA-1B, a model size where Mix-LN fails to converge. These findings demonstrate that \ours provides a robust and computationally efficient normalization strategy, enhancing large-scale training of language models without additional implementation complexity.

\subsection{Supervised Fine-tuning}
To verify whether the gains in pre-training can be translated to the stage of post-training, we perform SFT with the models obtained from Section \ref{sec:pre} on the Commonsense170K dataset \cite{hu2023llm} across eight downstream tasks. We adopt the same fine-tuning configurations as used in \citet{li2024mix}. The results, presented in Table~\ref{tab:model_comparison}, demonstrate that \ours consistently surpasses other normalization techniques in all evaluated datasets. For the LLaMA-250M model, \ours improves average performance by 1.80\% and achieves a 3.56\% gain on ARC-e compared to Mix-LN. Similar trends are observed with the LLaMA-1B model, where \ours outperforms Pre-LN, Post-LN, Mix-LN, and DeepNorm on seven out of eight tasks,  with an average gain of 1.86\% over the best baseline. These results confirm that \ours enhances generalization on diverse downstream tasks by improving the representation quality of deep layers.

\begin{table*}[t]
\centering
\caption{Fine-tuning performance ($\uparrow$) of LLaMA with various layer normalizations.}
    \resizebox{0.95\textwidth}{!}{%
\begin{tabular}{lcccccccc}
\toprule
\textbf{Method} & \textbf{MMLU} & \textbf{BoolQ} & \textbf{ARC-e} & \textbf{PIQA} & \textbf{Hellaswag} & \textbf{OBQA} & \textbf{Winogrande} & \textbf{Average} \\ 
\midrule
\multicolumn{9}{c}{\textbf{LLaMA-250M}} \\
Post-LN \citep{ba2016layer} & 22.95 & 37.83 & 26.94 & 52.72 & 26.17 & 11.60 & 49.56 & 32.54 \\
DeepNorm \citep{wang2024deepnet}& 23.60 & 37.86 & 36.62 & 61.10 & 25.69 & 15.00 & 49.57 & 35.63 \\

Mix-LN \citep {li2024mix}& 26.53 & 56.12 & 41.68 & 66.34 & 30.16 & 18.00 & 50.56 & 41.34 \\

Pre-LN \citep{baevski2018adaptive} & 24.93 & 38.35 & 40.15 & 63.55 & 26.34 & 16.20 & 49.01  & 36.93 \\
\midrule

Pre-LN + LayerNorm Scaling & \textbf{27.08} & \textbf{58.17} & \textbf{45.24} & \textbf{67.38} & \textbf{32.81} & \textbf{18.80} & \textbf{52.49}  & \textbf{43.14} \\

\midrule
\multicolumn{9}{c}{\textbf{LLaMA-1B}} \\
Post-LN  \citep{ba2016layer}& 22.95 & 37.82 & 25.08 & 49.51 & 25.04 & 13.80 & 49.57 & 31.96 \\
DeepNorm \citep{wang2024deepnet}& 23.35 & 37.83 & 27.06 & 52.94 & 26.19 & 11.80 & 49.49 & 32.67 \\

Mix-LN \citep{li2024mix} & 23.19 & 37.83 & 25.08 & 49.51 & 25.04 & 11.80 & 49.57 & 31.72 \\

Pre-LN \citep{baevski2018adaptive} & 26.54 & \textbf{62.20} & 45.70 & 67.79 & 30.96 & 17.40 & 50.51 & 43.01\\
\midrule

Pre-LN + LayerNorm Scaling & \textbf{28.69} & 61.80 & \textbf{48.85} & \textbf{67.92} & \textbf{33.94} & \textbf{18.60} & \textbf{54.30} & \textbf{44.87} \\
\bottomrule
\end{tabular}}

\label{tab:model_comparison}
\end{table*}

\subsection{Scaling Up Training}

\subsubsection{OLMo} 

\textbf{Model Size Scaling.} To further assess the scalability and robustness of LNS, we conduct additional experiments using the OLMo repository \citep{groeneveld2024olmo}, scaling training across model sizes of 60M, 150M, 300M, 1B, and 7B parameters. All models are trained on a fixed 20B-token budget to ensure comparability. These experiments are designed to evaluate whether the performance gains observed with LNS in smaller-scale settings extend to more challenging and state-of-the-art LLM training regimes. As shown in Figure~\ref{fig:illus_scaling}, LNS consistently and substantially outperforms the standard Pre-LN baseline across all model sizes. Remarkably, for the 7B model, LNS reduces the final loss from 2.69 to 2.50. These results underscore the scalability of LNS and its effectiveness in large-scale pre-training scenarios.

\begin{figure*}[h]
  \centering
    \includegraphics[width=0.55\linewidth]{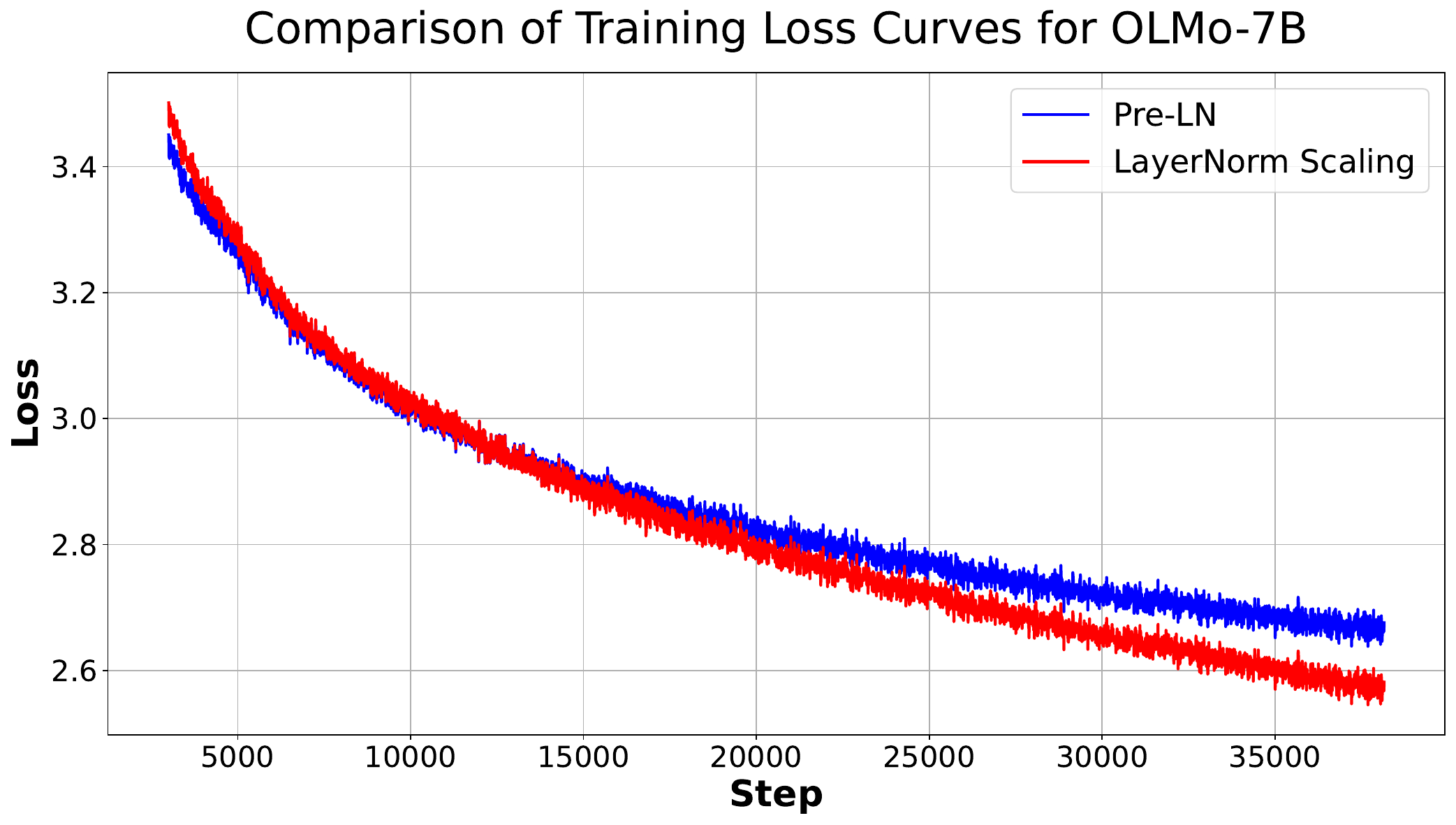}
  \caption{Training loss of OLMo-7B with Pre-LN and LNS.}
  \label{fig:train_loss_7b}
\end{figure*}

\textbf{Loss Curve.} Figure~\ref{fig:train_loss_7b} shows the training loss curves of 7B models trained with Pre-LN and LNS. While \ours exhibits slightly slower convergence at the early stages of training, it consistently outperforms Pre-LN as training progresses, ultimately achieving a substantial loss gap. We attribute this to the uncontrolled accumulation of output variance in Pre-LN, which amplifies with depth and training steps, ultimately impairing the effective learning of deeper layers. In contrast, LNS mitigates this issue by scaling down the output variance in proportion to depth, thereby enabling more stable and effective training across all layers during training.

\newcommand\dmodel{d_{\mathrm{model}}}

\textbf{Beating OLMo's Scaled Initialization.} OLMo adopts the scaled initialization proposed in~\citet{zhang2019improving} and used by~\citet{mehta2024openelm}, which scales input projections by $1 / \sqrt{\dmodel}$, and output projections by $1 / \sqrt{2 \cdot \dmodel \cdot l}$ at every layer. This method is designed to enhance training stability and to scale down variance at initialization. 
To evaluate the effectiveness of LNS, we compare it against this state-of-the-art initialization by training OLMo-1B on 20B tokens. As shown in Table~\ref{tab:olmo_init}, LNS achieves consistently lower training loss, indicating that it may offer a more effective alternative for large-scale LLM training. 

\begin{table}[h!]
\footnotesize
\centering
\caption{Comparison with OLMo's Scaled Initialization.}
\label{tab:scaling_pretrain}
\begin{tabular}{lcccc}
\toprule
Method & Model    & \# Tokens & Training Loss & Perplexity \\
\midrule
OLMo's Scaled Initialization & OLMo-1B  & 20B  & 2.96 &   19.30       \\
LayerNorm Scaling & OLMo-1B  & 20B  & \bf 2.85 &  \bf 17.28         \\
\bottomrule
\end{tabular}
\label{tab:olmo_init}
\end{table}


\subsubsection{Qwen2.5} We further evaluate the generalizability of LNS by applying it to a state-of-the-art architecture, Qwen2.5-0.5B \citep{qwen2.5}. We train the model for 6B tokens and compare LNS against the standard Pre-LN setup. Consistent with previous findings, Table \ref{tab:qwen} illustrates that LNS yields a notable reduction in perplexity—from 20.62 to 19.57—highlighting its effectiveness even on strong, modern architectures.

\begin{table}[h!]
\footnotesize
\centering
\caption{Perplexity (PPL $\downarrow$) comparison under scaled-up pre-training. For LLaMA-1B and 7B, training is scheduled for 100B tokens but is terminated early to report results. Qwen-2.5 is trained with a fixed budget of 6B tokens.}
\label{tab:scaling_pretrain}
\begin{tabular}{lcccc}
\toprule
Model         & \# Params & \# Tokens & Pre-LN (PPL) & LNS (PPL) \\
\midrule
Qwen2.5-0.5B  & 0.5B      & 6B        & 20.62        & \bf 19.57     \\
\bottomrule
\end{tabular}
\label{tab:qwen}
\end{table}

The consistent benefits observed across increased model scales, larger training datasets, and diverse architectures suggest that LNS is a promising technique for enhancing the training of contemporary large language models, ensuring that deeper layers contribute more effectively to learning.

\subsection{LNS Effectively Scales Down Output Variance}

As LNS is proposed to reduce output variance, we empirically validate this claim during the pre-training of LLMs. We compare the layerwise output variance of three configurations: (1) the standard Pre-LN~\citep{ba2016layer}, (2) Pre-LN with Scaled Initialization~\citep{shoeybi2019megatron,radford2019language}, which scales the initialization of the feedforward layers' weights $W_0$ and $W_2$ by $\frac{1}{\sqrt{2L}}$, where $L$ is the total number of Transformer layers, and (3) Pre-LN with LNS. The average output variance across layers is shown in Figure~\ref{fig:combined_variance}. For both vanilla Pre-LN and Scaled Initialization, the output variance in shallow layers (blue) remains relatively stable throughout training, while variance in deeper layers (red) grows substantially after 2K iterations, reaching up to 175 in the final layer. Since Scaled Initialization only operates at initialization, it is insufficient to constrain output variance during training. In contrast, LNS consistently suppresses the growth of output variance in deeper layers, capping it at approximately 25.

\subsection{LNS Enhances the Effectiveness of Deep Layers}

\begin{figure*}[!ht]
\centering

\begin{minipage}{0.49\textwidth}
    \centering
    \includegraphics[width=1.0\linewidth]{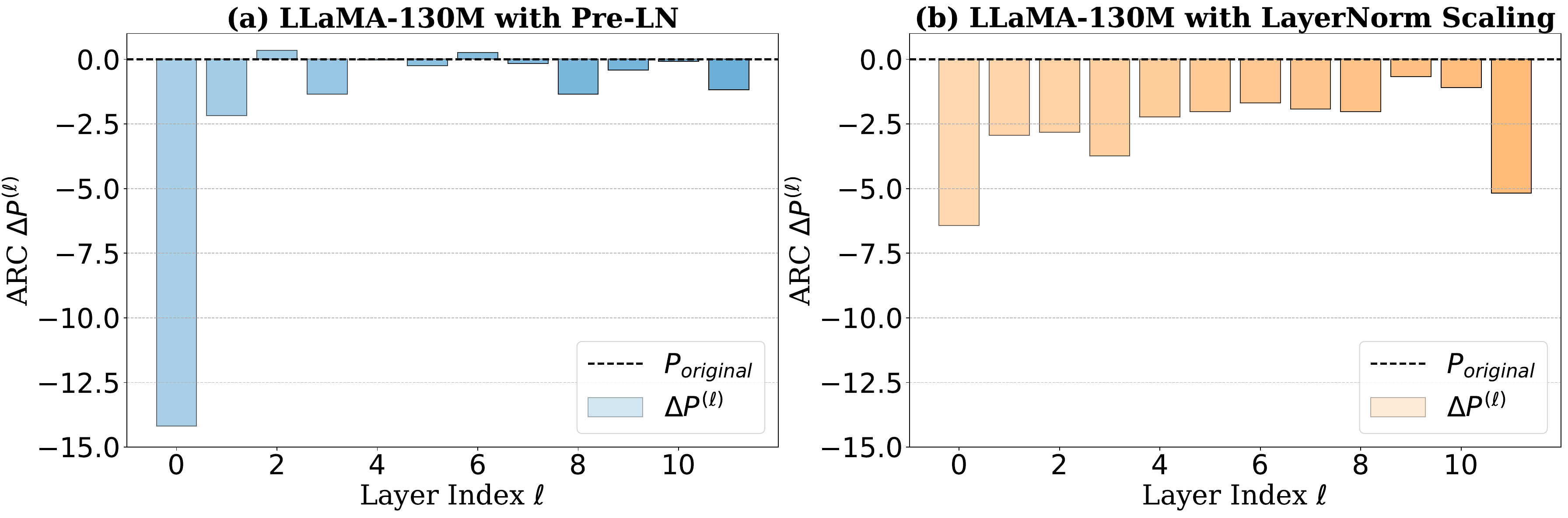}
\end{minipage}
\begin{minipage}{0.49\textwidth} 
    \centering
    \includegraphics[width=1.0\linewidth]{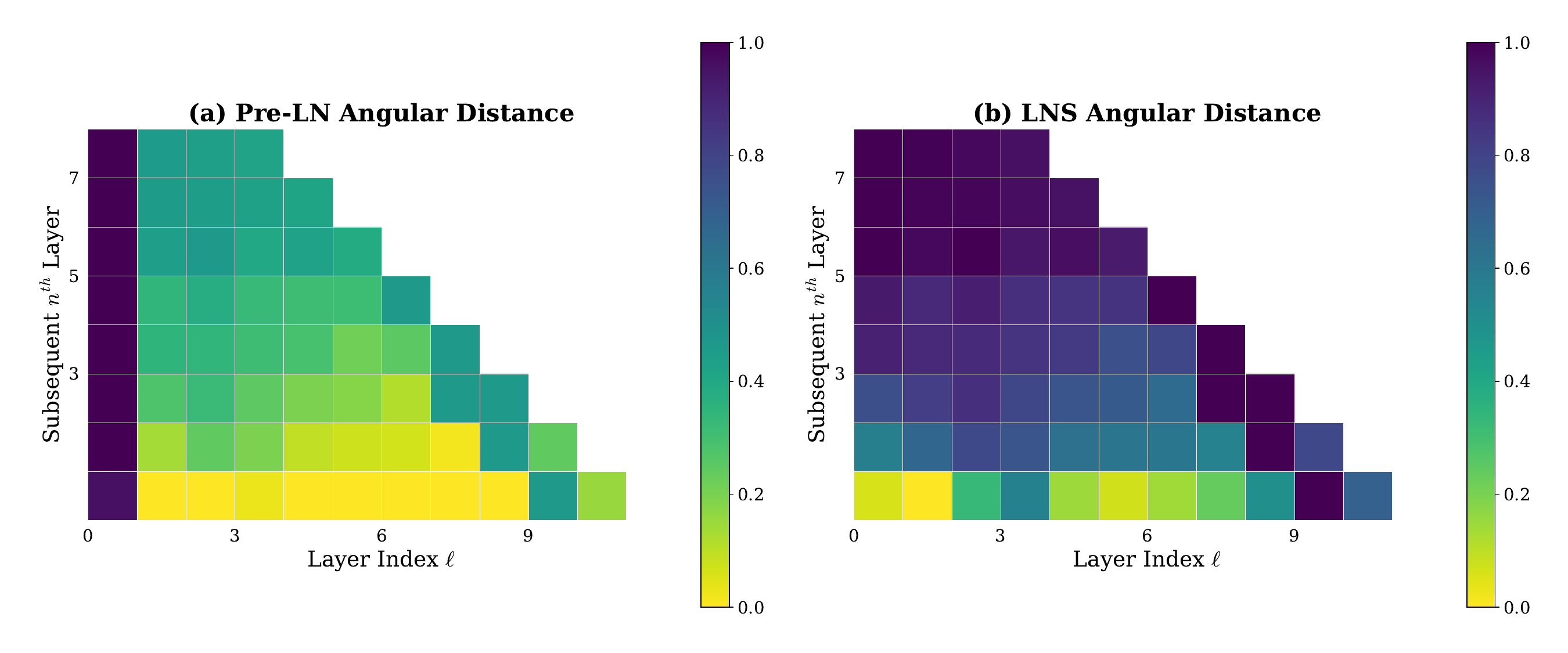}
\end{minipage}

\caption{
\textbf{Left:} Performance drop of layer pruning on LLaMA-130M. \textbf{Right:}
The angular distance between representations of subsequent layers is shown. LayerNorm Scaling enables deep layers to make a meaningful contribution to the model.
}

\label{fig:llama130_pruning}
\end{figure*}



Furthermore, to assess whether LNS enhances the effectiveness of deeper layers by promoting more diverse feature representations, we analyze the layerwise performance drop and the angular distance of LNS, as shown in Figure~\ref{fig:llama130_pruning}. Compared to Pre-LN, the performance degradation in \ours is more uniformly distributed across layers, indicating a more balanced contribution from each layer. Notably, pruning the deeper layers of LNS results in a more significant accuracy drop, suggesting these layers play a more critical role in task performance. Additionally, features learned under LNS exhibit greater distinction: most layers show a substantial angular distance, exceeding 0.6, from their adjacent layers, indicating more diverse representations. In sharp contrast, the layerwise angular distance in Pre-LN remains significantly lower and progressively decreases with depth, suggesting reduced feature diversity.


\subsection{LayerNorm Scaling in Vision Transformer}
\label{appendix:vision_transformer}

To evaluate whether LNS also benefits architectures beyond language models, we conduct experiments on ViT-S on ImageNet-1K. Since ViT-S includes LayerScale \citep{touvron2021going} by default---which may interfere with the effect of LNS---we remove LayerScale from all evaluated variants to ensure a fair comparison. We then test different insertion positions of LNS. The top-1 accuracy results are summarized in Table~\ref{tab:vit_lns}. Whereas LNS in language models is typically most effective directly after normalization, in Vision Transformers, the best position is after the attention and MLP blocks. We next examine whether this performance gain correlates with better control of layer-wise variance.

\begin{table}[!ht]
\centering
\caption{Top-1 accuracy (\%) of ViT-S model with and without LNS.}
\label{tab:vit_lns}
\begin{tabular}{lcc}
\toprule
\textbf{Model Variant} & \textbf{LNS Position} & \textbf{Top-1 Accuracy} \\
\midrule
ViT (w/o LayerScale) & -- & 67.91 \\
ViT (w/o LayerScale) & after LayerNorm & 66.43 \\
ViT (w/o LayerScale) & after Attn/MLP & \textbf{68.75} \\
\bottomrule
\end{tabular}
\end{table}

\begin{figure}[!ht]
    \centering
    \includegraphics[width=1.0\linewidth]{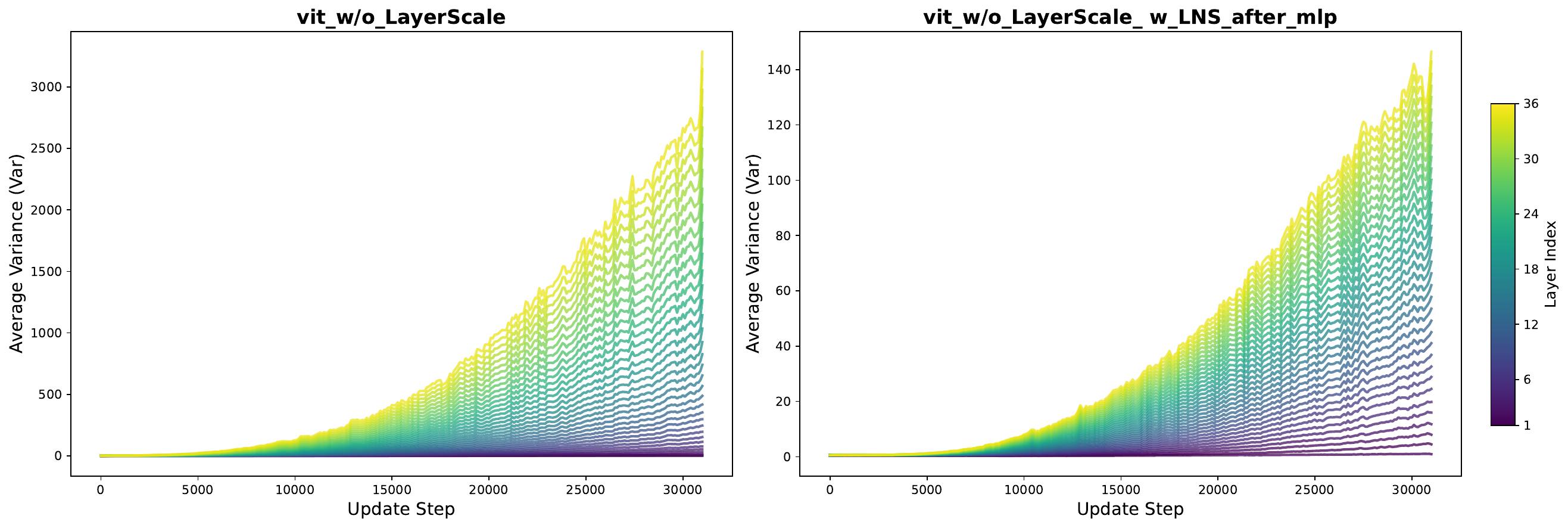}
    \caption{Layer-wise output variance of ViT-S \textbf{without} LayerScale (left) and with \textbf{LNS after Attn/MLP} (right). LNS significantly reduces the variance growth compared to the baseline.}
    \label{fig:vit_variance}
\end{figure}

Figure \ref{fig:vit_variance} plots the average output variance of each transformer block during training.  
Without LayerScale, variance in deeper layers grows rapidly—exceeding \(\sim\!3{,}000\) by 30K update steps.  
Applying LNS after Attn/MLP controls this growth to below \(\sim\!150\), confirming that LNS stabilizes the forward signal even in vision transformers.

These preliminary findings indicate that the variance-control mechanism underlying LNS generalizes to vision transformers when the scaling is applied after Attn/MLP.  
We leave a more detailed theoretical understanding of this behavior to future work and community discussion.

\section{Ablation Study}

\textbf{Comparing Against Other Scaling Methods.} We first compare LNS with previous scaling approaches, including (1) Scaled Initialization \citep{shoeybi2019megatron,radford2019language}, which scales the initialization of $W_0$ and $W_2$ by the overall depth $1/\sqrt{2L}$; (2) Depth-Scaled Initialization \citep{zhang2019improving} scales the initialization of weight matrices by the current depth $1/\sqrt{2l}$; (3) SkipInit \citep{de2020batchnormalizationbiasesresidual} introduces a learnable parameter after FFN/Att layers, initialized as $1/\sqrt{L}$; (4) LayerScale \citep{touvron2021going} applies per-channel weighting using a diagonal matrix, \textit{diag}$(\lambda_1, \ldots, \lambda_d)$, where each weight $\lambda_i$ is initialized to a small value (e.g., $\lambda_i = \epsilon$). Table~\ref{tab:scal_comparison} presents the results of LLaMA-130M and LLaMA-250M. 

First, we observe that methods involving learnable parameters, such as LayerScale and SkipInit, consistently degrade performance in LLMs. Among initialization-based techniques, a larger scaling factor proves beneficial: Scaled Initialization yields lower perplexity compared to Depth-Scaled Initialization. Notably, LNS achieves the best overall performance, underscoring the advantage of applying scaling dynamically during training. \textbf{Interestingly}, combining LNS with Scaled Initialization results in worse performance than using LNS alone, highlighting the importance of removing conflicting initialization strategies prior to adopting LNS.

\begin{table}[!ht]
\caption{Comparing LNS against other scaling methods. Perplexity (↓) is reported.  }
\vskip -0.3in
\label{tab:scal_comparison}
\begin{center}
\resizebox{0.8\columnwidth}{!}{%
\begin{tabular}{lcc}
\toprule
& \textbf{LLaMA-130M} & \textbf{LLaMA-250M} \\ 

Training Tokens & 2.2B & 3.9B \\
\midrule
Pre-LN & 26.73 & 21.92 \\
\midrule
+ LayerScale \citep{touvron2021going}  & 27.93 & 23.45 \\
+ SkipInit \citep{de2020batchnormalizationbiasesresidual} & 27.41 & 22.29 \\
+ Depth-Scaled Initialization \citep{zhang2019improving} & 26.95 & 21.50 \\
+ Scaled Initialization \citep{shoeybi2019megatron} & 26.04 & 20.98\\
+ LayerNorm Scaling & \textbf{25.76} & \textbf{20.35} \\
+ LayerNorm Scaling + Scaled Initialization   & 25.80 & 20.79 \\
\bottomrule

\end{tabular}}
\vskip -0.2in
\end{center}
\end{table}

\textbf{Comparison with Other Layer Normalization.} In addition, we conducted comparisons using LLaMA-130M to evaluate \ours against recently proposed normalization methods, including Admin~\citep{liu2020understanding}, Sandwich-LN~\citep{ding2021cogview}, Group-LN~\citep{wu2018group, ma2024megalodon}, and Mix-LN \citep{li2024mix}. Table \ref{tab:com_other} shows that Admin and Group-LN degrade performance. Sandwich-LN slightly outperforms Pre-LN. Both Mix-LN and \ours improve over Pre-LN by good margins. However, Mix-LN fails to reduce perplexity under 26, falling short of \ours and suffers from instability in large-scale scenarios as shown in Table \ref{tab:norm_comparison}.

\begin{table}[!ht]
\centering
\footnotesize
\caption{Comparison against other normalization methods on LLaMA-130M. All methods use the identical configurations. Perplexity (↓) is reported. }

\resizebox{0.8\columnwidth}{!}{%
\begin{tabular}{l c c c c c}
\toprule
 \textbf{Pre-LN} & \textbf{Admin} & \textbf{Group-LN} & \textbf{Sandwich-LN} & \textbf{Mix-LN} & \bf LayerNorm Scaling \\
\midrule
 26.73 & 27.91 & 28.01 & 26.51 & 26.07 &\textbf{25.76}  \\
\bottomrule
\end{tabular}}
\label{tab:com_other}
\end{table}

\textbf{Effect of Positions of LNS.}
The results in Table~\ref{tab:com_position} show that inserting the scaling factor at different points can have a considerable influence on the model’s performance. Placing it after the residual connection (“After Residual”) leads to a perplexity of 1358.11, which indicates training divergence. In contrast, LNS incorporates the scaling factor after LN achieving the best perplexity of 25.76, surpassing both the baseline Pre-LN setting (26.73) and other placements. This suggests that modifying the LayerNorm to include the scaling factor can enhance training stability and final performance for this model configuration.

\begin{table}[!ht]
\centering
\caption{Effects of Insertion Position of LayerNorm Scaling on LLaMA-130M}
\resizebox{0.95\columnwidth}{!}{%
\begin{tabular}{l| c c  c c c |c}
\toprule
 \textbf{Pre-LN} & \textbf{Before LN} & \textbf{After Attn/FFN} & \textbf{After Residual}  & \textbf{LNS Only After Attn}  & \textbf{LNS Only After FFN}  &  \bf Ours (After LN) \\
\midrule
 26.73 & 26.97 & 26.53 & 1358.11  & 26.89  & 26.43  &\textbf{25.76}  \\
\bottomrule
\end{tabular}}
\label{tab:com_position}
\end{table}

\vspace{-1em}
\section{Related Work}

\textbf{Ineffectiveness of Deeper Layers in Transformers.}  The ineffectiveness of deep layers in LLMs has been previously reported. \citet{yin2023outlier} found that deeper layers of LLMs can tolerate significantly higher levels of pruning compared to shallower layers, achieving high sparsity. Similarly, \citet{gromov2024unreasonable} and \citet{men2024shortgpt} demonstrated that removing early layers causes a dramatic decline in model performance, whereas removing deep layers does not. \citet{lad2024remarkable} showed that the middle and deep layers of GPT-2 and Pythia exhibit remarkable robustness to perturbations such as layer swapping and layer dropping. Recently, \citet{li2024owlore} highlighted that early layers contain more outliers and are therefore more critical for fine-tuning. While these studies effectively highlight the limitations of deep layers in LLMs, they stop short of identifying the root cause of this issue or proposing viable solutions to address it.

\noindent\textbf{Layer Normalization in Language Models.}
LN \citep{ba2016layer} was initially applied after the residual connection in the original Transformer \citep{vaswani2017attention}, which is known as Post-LN. Later on, Pre-LN  \citep{baevski2018adaptive,dai2019transformer,nguyen2019transformers} dominated LLMs, due to its compelling performance and stability \citep{brown2020language,touvron2023llama,jiang2023mistral,bi2024deepseek}. Prior works have studied the effect of Pre-LN and Post-LN. 
\citet{xiong2020layer} proves that Post-LN tends to have larger gradients near the output layer, which necessitates smaller learning rates to stabilize training, whereas Pre-LN scales down gradients with the depth of the model, working better for deep Transformers.  \citet{wang2019learning} empirically confirmed that Pre-LN facilitates stacking more layers and Post-LN suffers from gradient vanishing. The idea of connecting multiple layers was proposed in previous works \citep{bapna2018training,dou2018exploiting,wang2019learning}. Admin introduces additional parameters to control residual dependencies, stabilizing Post-LN. DeepNorm \citep{wang2024deepnet} enables stacking 1000-layer Transformers by upscaling the residual connection before applying LN. Additionally, \citet{ding2021cogview} proposed Sandwich LayerNorm, normalizing both the input and output of each transformer sub-layer. \citet{takase2022b2t} introduced B2T to bypass all LN except the final one in each layer. \citet{li2024mix} recently combines Post-LN and Pre-LN to enhance the middle layers. \citet{zhu2025transformersnormalization} introduces Dynamic Tanh (DyT) as a normalization-free alternative in Transformers, delivering comparable performance.  \citet{zhuo2025hybridnormstableefficienttransformer} proposes HybridNorm, a hybrid normalization scheme combining QKV normalization with Post-Norm FFN to stabilize training in deep transformers. \citet{de2020batchnormalizationbiasesresidual} also states that normalized residual blocks in deep networks are close to the identity function and proposes SkipInit to remove normalization by introducing a learnable scalar multiplier on the residual branch initialized to $1/\sqrt{L}$. Our experiments suggest that SkipInit's learnable parameter does not improve performance and sometimes harms training. 

\noindent\textbf{Depth Scaling, Parameterization, and Hyperparameter Transfer.}
The $\mu$P framework \citep{yang2021tuning} enables hyperparameters tuned on small-width proxy models to transfer to larger-width models while preserving feature learning.
However, width-wise $\mu$P does not fully resolve the challenges introduced by depth scaling.
Depth-$\mu$P \citep{yang2024tensor} extends this perspective to infinite-depth networks by scaling both residual-branch contributions and parameter updates by $1/\sqrt{L}$, which for Adam-type optimizers entails depth-dependent learning-rate scaling.
Similarly, \citet{bordelon2024depthwise} combine $\mu$P with $1/\sqrt{L}$ residual scaling to achieve hyperparameter transfer across width and depth in residual networks and Vision Transformers.
CompleteP \citep{dey2026don} further argues that useful depth parameterizations should support both depth-wise hyperparameter transfer and non-lazy feature learning in all layers, while avoiding the need to rescale the base learning rate with depth.
\citet{mlodozeniec2025completed} extend this direction to transfer across modules, width, depth, batch size, and training duration. Several theoretical works further analyze stability and degeneracy in very deep networks.
Stable ResNet \citep{hayou2021stable}, the neural covariance SDE \citep{li2022neural}, differential-equation scaling limits \citep{li2023differential}, and the Shaped Transformer \citep{noci2023shaped} study residual scaling, covariance dynamics, and signal propagation in infinite-depth or infinite-depth-and-width regimes.
For Transformers, \citet{noci2022signal} show that self-attention layers can suffer from rank collapse and that depth-dependent residual scaling can mitigate signal degeneration.
\citet{sanyal2024attention} observe attention collapse in deeper layers of decoder-only LLMs and exploit this degeneracy for model compression. 
However, prior works mainly derive parameterization or optimizer-scaling rules that make training dynamics and hyperparameters transferable across model scales, whereas we study why deeper layers in standard Pre-LN LLMs become functionally ineffective. 
We identify variance growth under Pre-LN as a mechanism that makes deeper Transformer blocks approach identity mappings, and propose LNS as a simple modification to LayerNorm to counteract this effect. 
Rather than providing a global recipe for hyperparameter transfer across model families, LNS directly improves the functional effectiveness of deep layers under otherwise standard training settings, without requiring the depth-dependent learning-rate scaling used in some depth-wise transfer methods.

\section{Conclusion}

In this paper, we re-introduce the concept of the \textit{Curse of Depth} in LLMs, highlighting an urgent yet often overlooked phenomenon: nearly half of the deep layers in modern LLMs are less effective than expected. We discover the root cause of this phenomenon is Pre-LN which is widely used in almost all modern LLMs. To tackle this issue, we introduce \textit{LayerNorm Scaling}. By scaling the output variance inversely with the layer depth, LayerNorm Scaling ensures that all layers, including deeper ones, contribute meaningfully to training. Our experiments show that this simple modification improves performance, reduces resource usage, and stabilizes training across various model sizes. LayerNorm Scaling is easy to implement, hyperparameter-free, and provides a robust solution to enhance the efficiency and effectiveness of LLMs.

\bibliographystyle{plainnat}
\bibliography{custom}

@misc{zhuo2025hybridnormstableefficienttransformer,
      title={HybridNorm: Towards Stable and Efficient Transformer Training via Hybrid Normalization}, 
      author={Zhijian Zhuo and Yutao Zeng and Ya Wang and Sijun Zhang and Jian Yang and Xiaoqing Li and Xun Zhou and Jinwen Ma},
      year={2025},
      eprint={2503.04598},
      archivePrefix={arXiv},
      primaryClass={cs.CL},
      url={https://arxiv.org/abs/2503.04598}, 
}

@misc{de2020batchnormalizationbiasesresidual,
      title={Batch Normalization Biases Residual Blocks Towards the Identity Function in Deep Networks}, 
      author={Soham De and Samuel L. Smith},
      year={2020},
      eprint={2002.10444},
      archivePrefix={arXiv},
      primaryClass={cs.LG},
      url={https://arxiv.org/abs/2002.10444}, 
}

@article{huang2025stable,
  title={Stable-SPAM: How to Train in 4-Bit More Stably than 16-Bit Adam},
  author={Huang, Tianjin and Hu, Haotian and Zhang, Zhenyu and Jin, Gaojie and Li, Xiang and Shen, Li and Chen, Tianlong and Liu, Lu and Wen, Qingsong and Wang, Zhangyang and others},
  journal={arXiv preprint arXiv:2502.17055},
  year={2025}
}

@article{huang2025spam,
  title={SPAM: Spike-Aware Adam with Momentum Reset for Stable LLM Training},
  author={Huang, Tianjin and Zhu, Ziquan and Jin, Gaojie and Liu, Lu and Wang, Zhangyang and Liu, Shiwei},
  journal={arXiv preprint arXiv:2501.06842},
  year={2025}
}

@article{zhu2025transformers,
  title={Transformers without normalization},
  author={Zhu, Jiachen and Chen, Xinlei and He, Kaiming and LeCun, Yann and Liu, Zhuang},
  journal={arXiv preprint arXiv:2503.10622},
  year={2025}
}

@article{du2021glm,
  title={Glm: General language model pretraining with autoregressive blank infilling},
  author={Du, Zhengxiao and Qian, Yujie and Liu, Xiao and Ding, Ming and Qiu, Jiezhong and Yang, Zhilin and Tang, Jie},
  journal={arXiv preprint arXiv:2103.10360},
  year={2021}
}

@misc{zhu2025transformersnormalization,
      title={Transformers without Normalization}, 
      author={Jiachen Zhu and Xinlei Chen and Kaiming He and Yann LeCun and Zhuang Liu},
      year={2025},
      eprint={2503.10622},
      archivePrefix={arXiv},
      primaryClass={cs.LG},
      url={https://arxiv.org/abs/2503.10622}, 
}

@article{men2024shortgpt,
  title={Shortgpt: Layers in large language models are more redundant than you expect},
  author={Men, Xin and Xu, Mingyu and Zhang, Qingyu and Wang, Bingning and Lin, Hongyu and Lu, Yaojie and Han, Xianpei and Chen, Weipeng},
  journal={arXiv preprint arXiv:2403.03853},
  year={2024}
}

@book{ledoux2001concentration,
  title={The Concentration of Measure Phenomenon},
  author={Ledoux, Michel},
  year={2001},
  publisher={American Mathematical Society},
  series={Mathematical Surveys and Monographs},
  volume={89},
  address={Providence, RI}
}

@article{jiang2023mistral,
  title={Mistral 7B},
  author={Jiang, Albert Q and Sablayrolles, Alexandre and Mensch, Arthur and Bamford, Chris and Chaplot, Devendra Singh and Casas, Diego de las and Bressand, Florian and Lengyel, Gianna and Lample, Guillaume and Saulnier, Lucile and others},
  journal={arXiv preprint arXiv:2310.06825},
  year={2023}
}

@article{yang2023tensor,
  title={Tensor programs vi: Feature learning in infinite-depth neural networks},
  author={Yang, Greg and Yu, Dingli and Zhu, Chen and Hayou, Soufiane},
  journal={arXiv preprint arXiv:2310.02244},
  year={2023}
}

@inproceedings{lialin2023relora,
  title={Relora: High-rank training through low-rank updates},
  author={Lialin, Vladislav and Muckatira, Sherin and Shivagunde, Namrata and Rumshisky, Anna},
  booktitle={ICLR},
  year={2023}
}

@article{hu2023llm,
  title={LLM-Adapters: An adapter family for parameter-efficient fine-tuning of large language models},
  author={Hu, Zhiqiang and Wang, Lei and Lan, Yihuai and Xu, Wanyu and Lim, Ee-Peng and Bing, Lidong and Xu, Xing and Poria, Soujanya and Lee, Roy Ka-Wei},
  journal={EMNLP},
  year={2023}
}

@article{kingma2014adam,
  title={Adam: A method for stochastic optimization},
  author={Kingma, Diederik P},
  journal={ICLR},
  year={2015}
}

@article{shazeer2020glu,
  title={Glu variants improve transformer},
  author={Shazeer, Noam},
  journal={arXiv preprint arXiv:2002.05202},
  year={2020}
}

@article{zhang2019root,
  title={Root mean square layer normalization},
  author={Zhang, Biao and Sennrich, Rico},
  journal={NeurIPS},
  volume={32},
  year={2019}
}

@inproceedings{touvron2021going,
  title={Going deeper with image transformers},
  author={Touvron, Hugo and Cord, Matthieu and Sablayrolles, Alexandre and Synnaeve, Gabriel and J{\'e}gou, Herv{\'e}},
  booktitle={ICCV},
  pages={32--42},
  year={2021}
}

@article{rajpurkar2016squad,
  title={Squad: 100,000+ questions for machine comprehension of text},
  author={Rajpurkar, P},
  journal={EMNLP},
  year={2016}
}

@article{hendrycks2020measuring,
  title={Measuring massive multitask language understanding},
  author={Hendrycks, Dan and Burns, Collin and Basart, Steven and Zou, Andy and Mazeika, Mantas and Song, Dawn and Steinhardt, Jacob},
  journal={ICLR},
  year={2021}
}

@article{baevski2018adaptive,
  title={Adaptive input representations for neural language modeling},
  author={Baevski, Alexei and Auli, Michael},
  journal={ICLR},
  year={2019}
}

@article{vaswani2017attention,
  title={Attention is all you need},
  author={Vaswani, A},
  journal={NeurIPS},
  year={2017}
}

@article{devlin2018bert,
  title={Bert: Pre-training of deep bidirectional transformers for language understanding},
  author={Devlin, Jacob},
  journal={NAACL},
  year={2019}
}

@inproceedings{xiong2020layer,
  title={On layer normalization in the transformer architecture},
  author={Xiong, Ruibin and Yang, Yunchang and He, Di and Zheng, Kai and Zheng, Shuxin and Xing, Chen and Zhang, Huishuai and Lan, Yanyan and Wang, Liwei and Liu, Tieyan},
  booktitle={ICML},
  pages={10524--10533},
  year={2020},
  organization={PMLR}
}

@article{ba2016layer,
  title={Layer normalization},
  author={Ba, Jimmy Lei},
  journal={arXiv preprint arXiv:1607.06450},
  year={2016}
}

@article{dai2019transformer,
  title={Transformer-xl: Attentive language models beyond a fixed-length context},
  author={Dai, Zihang and Yang, Zhilin and Yang, Yiming and Carbonell, Jaime and Le, Quoc V and Salakhutdinov, Ruslan},
  journal={ACL},
  year={2019}

}

@article{nguyen2019transformers,
  title={Transformers without tears: Improving the normalization of self-attention},
  author={Nguyen, Toan Q and Salazar, Julian},
  journal={IWSLT},
  year={2019}
}

@article{ding2021cogview,
  title={Cogview: Mastering text-to-image generation via transformers},
  author={Ding, Ming and Yang, Zhuoyi and Hong, Wenyi and Zheng, Wendi and Zhou, Chang and Yin, Da and Lin, Junyang and Zou, Xu and Shao, Zhou and Yang, Hongxia and others},
  journal={NeurIPS},
  volume={34},
  pages={19822--19835},
  year={2021}
}

@article{ma2024megalodon,
  title={Megalodon: Efficient llm pretraining and inference with unlimited context length},
  author={Ma, Xuezhe and Yang, Xiaomeng and Xiong, Wenhan and Chen, Beidi and Yu, Lili and Zhang, Hao and May, Jonathan and Zettlemoyer, Luke and Levy, Omer and Zhou, Chunting},
  journal={NeurIPS},
  year={2024}
}

@inproceedings{wu2018group,
  title={Group normalization},
  author={Wu, Yuxin and He, Kaiming},
  booktitle={ECCV},
  pages={3--19},
  year={2018}
}

@article{takase2023spike,
  title={Spike No More: Stabilizing the Pre-training of Large Language Models},
  author={Takase, Sho and Kiyono, Shun and Kobayashi, Sosuke and Suzuki, Jun},
  journal={arXiv preprint arXiv:2312.16903},
  year={2023}
}

@article{lad2024remarkable,
  title={The Remarkable Robustness of LLMs: Stages of Inference?},
  author={Lad, Vedang and Gurnee, Wes and Tegmark, Max},
  journal={arXiv preprint arXiv:2406.19384},
  year={2024}
}

@article{dou2018exploiting,
  title={Exploiting deep representations for neural machine translation},
  author={Dou, Zi-Yi and Tu, Zhaopeng and Wang, Xing and Shi, Shuming and Zhang, Tong},
  journal={EMNLP},
  year={2018}
}

@article{bapna2018training,
  title={Training deeper neural machine translation models with transparent attention},
  author={Bapna, Ankur and Chen, Mia Xu and Firat, Orhan and Cao, Yuan and Wu, Yonghui},
  journal={EMNLP},
  year={2018}
}

@article{wang2019learning,
  title={Learning deep transformer models for machine translation},
  author={Wang, Qiang and Li, Bei and Xiao, Tong and Zhu, Jingbo and Li, Changliang and Wong, Derek F and Chao, Lidia S},
  journal={ACL},
  year={2019}
}

@article{siddiqui2024deeper,
  title={A deeper look at depth pruning of LLMs},
  author={Siddiqui, Shoaib Ahmed and Dong, Xin and Heinrich, Greg and Breuel, Thomas and Kautz, Jan and Krueger, David and Molchanov, Pavlo},
  journal={ICML},
  year={2024}
}

@article{dumitru2024layer,
  title={Layer-Wise Quantization: A Pragmatic and Effective Method for Quantizing LLMs Beyond Integer Bit-Levels},
  author={Dumitru, Razvan-Gabriel and Yadav, Vikas and Maheshwary, Rishabh and Clotan, Paul-Ioan and Madhusudhan, Sathwik Tejaswi and Surdeanu, Mihai},
  journal={arXiv preprint arXiv:2406.17415},
  year={2024}
}

@article{shoeybi2019megatron,
  title={Megatron-lm: Training multi-billion parameter language models using model parallelism},
  author={Shoeybi, Mohammad and Patwary, Mostofa and Puri, Raul and LeGresley, Patrick and Casper, Jared and Catanzaro, Bryan},
  journal={ICML},
  year={2020}
}

@inproceedings{muralidharan2024compact,
  title={Compact language models via pruning and knowledge distillation},
  author={Muralidharan, Saurav and Sreenivas, Sharath Turuvekere and Joshi, Raviraj Bhuminand and Chochowski, Marcin and Patwary, Mostofa and Shoeybi, Mohammad and Catanzaro, Bryan and Kautz, Jan and Molchanov, Pavlo},
  booktitle={NeurIPS},
  year={2024}
}

@article{lu2024alphapruning,
  title={Alphapruning: Using heavy-tailed self regularization theory for improved layer-wise pruning of large language models},
  author={Lu, Haiquan and Zhou, Yefan and Liu, Shiwei and Wang, Zhangyang and Mahoney, Michael W and Yang, Yaoqing},
  journal={NeurIPS},
  year={2024}
}

@article{li2024owlore,
  title={OwLore: Outlier-weighed Layerwise Sampled Low-Rank Projection for Memory-Efficient LLM Fine-tuning},
  author={Li, Pengxiang and Yin, Lu and Gao, Xiaowei and Liu, Shiwei},
  journal={arXiv preprint arXiv:2405.18380},
  year={2024}
}

@article{gromov2024unreasonable,
  title={The unreasonable ineffectiveness of the deeper layers},
  author={Gromov, Andrey and Tirumala, Kushal and Shapourian, Hassan and Glorioso, Paolo and Roberts, Daniel A},
  journal={arXiv preprint arXiv:2403.17887},
  year={2024}
}

@article{takase2022b2t,
  title={B2t connection: Serving stability and performance in deep transformers},
  author={Takase, Sho and Kiyono, Shun and Kobayashi, Sosuke and Suzuki, Jun},
  journal={ACL},
  year={2023}
}

@article{liu2020understanding,
  title={Understanding the difficulty of training transformers},
  author={Liu, Liyuan and Liu, Xiaodong and Gao, Jianfeng and Chen, Weizhu and Han, Jiawei},
  journal={EMNLP},
  year={2020}
}

@article{zhao2024galore,
  title={Galore: Memory-efficient llm training by gradient low-rank projection},
  author={Zhao, Jiawei and Zhang, Zhenyu and Chen, Beidi and Wang, Zhangyang and Anandkumar, Anima and Tian, Yuandong},
  journal={ICML},
  year={2024}
}

@article{bi2024deepseek,
  title={Deepseek llm: Scaling open-source language models with longtermism},
  author={Bi, Xiao and Chen, Deli and Chen, Guanting and Chen, Shanhuang and Dai, Damai and Deng, Chengqi and Ding, Honghui and Dong, Kai and Du, Qiushi and Fu, Zhe and others},
  journal={arXiv preprint arXiv:2401.02954},
  year={2024}
}

@article{wang2024deepnet,
  title={Deepnet: Scaling transformers to 1,000 layers},
  author={Wang, Hongyu and Ma, Shuming and Dong, Li and Huang, Shaohan and Zhang, Dongdong and Wei, Furu},
  journal={TPAMI},
  year={2024},
  publisher={IEEE}
}

@article{yin2023outlier,
  title={Outlier weighed layerwise sparsity (owl): A missing secret sauce for pruning llms to high sparsity},
  author={Yin, Lu and Wu, You and Zhang, Zhenyu and Hsieh, Cheng-Yu and Wang, Yaqing and Jia, Yiling and Pechenizkiy, Mykola and Liang, Yi and Wang, Zhangyang and Liu, Shiwei},
  journal={ICML},
  year={2024}
}

@article{touvron2023llama,
  title={Llama: Open and efficient foundation language models},
  author={Touvron, Hugo and Lavril, Thibaut and Izacard, Gautier and Martinet, Xavier and Lachaux, Marie-Anne and Lacroix, Timoth{\'e}e and Rozi{\`e}re, Baptiste and Goyal, Naman and Hambro, Eric and Azhar, Faisal and others},
  journal={arXiv preprint arXiv:2302.13971},
  year={2023}
}

@article{brown2020language,
  title={Language models are few-shot learners},
  author={Brown, Tom and Mann, Benjamin and Ryder, Nick and Subbiah, Melanie and Kaplan, Jared D and Dhariwal, Prafulla and Neelakantan, Arvind and Shyam, Pranav and Sastry, Girish and Askell, Amanda and others},  journal={NeurIPS},
  year={2020}
}

@inproceedings{yue2024mmmu,
  title={Mmmu: A massive multi-discipline multimodal understanding and reasoning benchmark for expert agi},
  author={Yue, Xiang and Ni, Yuansheng and Zhang, Kai and Zheng, Tianyu and Liu, Ruoqi and Zhang, Ge and Stevens, Samuel and Jiang, Dongfu and Ren, Weiming and Sun, Yuxuan and others},
  booktitle={Proceedings of the IEEE/CVF Conference on Computer Vision and Pattern Recognition},
  pages={9556--9567},
  year={2024}
}

@article{bai2025qwen2,
  title={Qwen2. 5-vl technical report},
  author={Bai, Shuai and Chen, Keqin and Liu, Xuejing and Wang, Jialin and Ge, Wenbin and Song, Sibo and Dang, Kai and Wang, Peng and Wang, Shijie and Tang, Jun and others},
  journal={arXiv preprint arXiv:2502.13923},
  year={2025}
}

@article{qwen2.5,
    title   = {Qwen2.5 Technical Report}, 
    author  = {An Yang and Baosong Yang and Beichen Zhang and Binyuan Hui and Bo Zheng and Bowen Yu and Chengyuan Li and Dayiheng Liu and Fei Huang and Haoran Wei and Huan Lin and Jian Yang and Jianhong Tu and Jianwei Zhang and Jianxin Yang and Jiaxi Yang and Jingren Zhou and Junyang Lin and Kai Dang and Keming Lu and Keqin Bao and Kexin Yang and Le Yu and Mei Li and Mingfeng Xue and Pei Zhang and Qin Zhu and Rui Men and Runji Lin and Tianhao Li and Tingyu Xia and Xingzhang Ren and Xuancheng Ren and Yang Fan and Yang Su and Yichang Zhang and Yu Wan and Yuqiong Liu and Zeyu Cui and Zhenru Zhang and Zihan Qiu},
    journal = {arXiv preprint arXiv:2412.15115},
    year    = {2024}
}

@article{radford2019language,
  title={Language models are unsupervised multitask learners},
  author={Radford, Alec and Wu, Jeffrey and Child, Rewon and Luan, David and Amodei, Dario and Sutskever, Ilya and others},
  journal={OpenAI blog},
  volume={1},
  number={8},
  pages={9},
  year={2019}
}

@online{qwen3_blog,
  author       = {Qwen Team},
  title        = {Qwen3: Think Deeper, Act Faster},
  year         = {2025},
  month        = {April},
  day          = {29},
  url          = {https://qwenlm.github.io/blog/qwen3/},
  note         = {Accessed: 2025-05-11}
}

@article{mehta2024openelm,
  title={Openelm: An efficient language model family with open training and inference framework},
  author={Mehta, Sachin and Sekhavat, Mohammad Hossein and Cao, Qingqing and Horton, Maxwell and Jin, Yanzi and Sun, Chenfan and Mirzadeh, Iman and Najibi, Mahyar and Belenko, Dmitry and Zatloukal, Peter and others},
  journal={arXiv preprint arXiv:2404.14619},
  year={2024}
}

@article{zhang2019improving,
  title={Improving deep transformer with depth-scaled initialization and merged attention},
  author={Zhang, Biao and Titov, Ivan and Sennrich, Rico},
  journal={arXiv preprint arXiv:1908.11365},
  year={2019}
}

@article{achiam2023gpt,
  title={Gpt-4 technical report},
  author={Achiam, Josh and Adler, Steven and Agarwal, Sandhini and Ahmad, Lama and Akkaya, Ilge and Aleman, Florencia Leoni and Almeida, Diogo and Altenschmidt, Janko and Altman, Sam and Anadkat, Shyamal and others},
  journal={arXiv preprint arXiv:2303.08774},
  year={2023}
}

@article{li2024mix,
  title={Mix-LN: Unleashing the Power of Deeper Layers by Combining Pre-LN and Post-LN},
  author={Li, Pengxiang and Yin, Lu and Liu, Shiwei},
  journal={arXiv preprint arXiv:2412.13795},
  year={2024}
}

@book{vershynin2018, 
place={Cambridge}, 
series={Cambridge Series in Statistical and Probabilistic Mathematics}, 
title={High-Dimensional Probability: An Introduction with Applications in Data Science}, 
publisher={Cambridge University Press}, author={Vershynin, Roman}, year={2018}, 
collection={Cambridge Series in Statistical and Probabilistic Mathematics}}

@article{groeneveld2024olmo,
  title={Olmo: Accelerating the science of language models},
  author={Groeneveld, Dirk and Beltagy, Iz and Walsh, Pete and Bhagia, Akshita and Kinney, Rodney and Tafjord, Oyvind and Jha, Ananya Harsh and Ivison, Hamish and Magnusson, Ian and Wang, Yizhong and others},
  journal={arXiv preprint arXiv:2402.00838},
  year={2024}
}

@book{Whittaker_Watson_1996, 
place={Cambridge}, 
edition={4}, 
series={Cambridge Mathematical Library}, 
title={A Course of Modern Analysis}, 
publisher={Cambridge University Press}, 
author={Whittaker, E. T. and Watson, G. N.}, 
year={1996}, 
collection={Cambridge Mathematical Library}}

@article{li2023differential,
  title={Differential equation scaling limits of shaped and unshaped neural networks},
  author={Li, Mufan Bill and Nica, Mihai},
  journal={arXiv preprint arXiv:2310.12079},
  year={2023}
}

@article{noci2023shaped,
  title={The shaped transformer: Attention models in the infinite depth-and-width limit},
  author={Noci, Lorenzo and Li, Chuning and Li, Mufan and He, Bobby and Hofmann, Thomas and Maddison, Chris J and Roy, Dan},
  journal={Advances in Neural Information Processing Systems},
  volume={36},
  pages={54250--54281},
  year={2023}
}

@article{mlodozeniec2025completed,
  title={Completed Hyperparameter Transfer across Modules, Width, Depth, Batch and Duration},
  author={Mlodozeniec, Bruno and Ablin, Pierre and B{\'e}thune, Louis and Busbridge, Dan and Klein, Michal and Ramapuram, Jason and Cuturi, Marco},
  journal={arXiv preprint arXiv:2512.22382},
  year={2025}
}

@article{dey2026don,
  title={Don't be lazy: CompleteP enables compute-efficient deep transformers},
  author={Dey, Nolan and Zhang, Bin and Noci, Lorenzo and Li, Mufan and Bordelon, Blake and Bergsma, Shane and Pehlevan, Cengiz and Hanin, Boris and Hestness, Joel},
  journal={Advances in Neural Information Processing Systems},
  volume={38},
  pages={137707--137739},
  year={2026}
}

@inproceedings{yang2024tensor,
  title={Tensor programs VI: Feature learning in infinite depth neural networks},
  author={Yang, Greg and Yu, Dingli and Zhu, Chen and Hayou, Soufiane},
  booktitle={International Conference on Learning Representations},
  volume={2024},
  pages={55099--55150},
  year={2024}
}

@inproceedings{bordelon2024depthwise,
  title={Depthwise hyperparameter transfer in residual networks: Dynamics and scaling limit},
  author={Bordelon, Blake and Noci, Lorenzo and Li, Mufan and Hanin, Boris and Pehlevan, Cengiz},
  booktitle={International Conference on Learning Representations},
  volume={2024},
  pages={22088--22127},
  year={2024}
}

@article{noci2022signal,
  title={Signal propagation in transformers: Theoretical perspectives and the role of rank collapse},
  author={Noci, Lorenzo and Anagnostidis, Sotiris and Biggio, Luca and Orvieto, Antonio and Singh, Sidak Pal and Lucchi, Aurelien},
  journal={Advances in Neural Information Processing Systems},
  volume={35},
  pages={27198--27211},
  year={2022}
}

@inproceedings{hayou2021stable,
  title={Stable resnet},
  author={Hayou, Soufiane and Clerico, Eugenio and He, Bobby and Deligiannidis, George and Doucet, Arnaud and Rousseau, Judith},
  booktitle={International Conference on Artificial Intelligence and Statistics},
  pages={1324--1332},
  year={2021},
  organization={PMLR}
}

@article{yang2021tuning,
  title={Tuning large neural networks via zero-shot hyperparameter transfer},
  author={Yang, Ge and Hu, Edward and Babuschkin, Igor and Sidor, Szymon and Liu, Xiaodong and Farhi, David and Ryder, Nick and Pachocki, Jakub and Chen, Weizhu and Gao, Jianfeng},
  journal={Advances in Neural Information Processing Systems},
  volume={34},
  pages={17084--17097},
  year={2021}
}

@article{li2022neural,
  title={The neural covariance SDE: Shaped infinite depth-and-width networks at initialization},
  author={Li, Mufan and Nica, Mihai and Roy, Dan},
  journal={Advances in Neural Information Processing Systems},
  volume={35},
  pages={10795--10808},
  year={2022}
}

@article{sanyal2024attention,
  title={When Attention Collapses: How Degenerate Layers in LLMs Enable Smaller, Stronger Models},
  author={Sanyal, Sunny and Shwartz-Ziv, Ravid and Dimakis, Alexandros G and Sanghavi, Sujay},
  journal={arXiv preprint arXiv:2404.08634},
  year={2024}
}

\newpage
\appendix
\onecolumn
\section{Proofs of the Theorems of curse of depth}
\label{appendix:proofs}
\subsection{Proof of Lemma \ref{attention_variance}}
\label{appendix:proof1}
\begin{proof}

Given Equation \eqref{x_result} from \cite{takase2023spike}, we have:
\begin{equation}
\begin{aligned}
    y &= x_{\ell+1} = x^\prime_\ell + \mathrm{FFN}(\mathrm{LN}(x^\prime_\ell)),  \\
    x^\prime_\ell &= x_\ell + \mathrm{Attn}(\mathrm{LN}(x_\ell)). 
\end{aligned}
\end{equation}

Based on our Assumption \ref{total_assumption}, let $ \mathrm{Var}(\mathrm{Attn}(\mathrm{LN}(x_\ell))) = \sigma_{\text{Attn}}^2 $. Then we can write:
\begin{equation}
\begin{aligned}
\mathrm{Var}(x^\prime_\ell) &= \mathrm{Var}(x_\ell) + \mathrm{Var}(\mathrm{Attn}(\mathrm{LN}(x_\ell))) + \mathrm{Cov}(\mathrm{Attn}(\mathrm{LN}(x_\ell)),\mathrm{Var}(x_\ell)) \\
&=\sigma_{x_\ell}^2 + \sigma_{\text{Attn}}^2 + \rho_1 \cdot \sigma_{x_\ell} \cdot  \sigma_{\text{Attn}},
\end{aligned}
\end{equation}

where $\rho_1$ is the correlation factor. Similarly, let $ \mathrm{Var}(\mathrm{FFN}(\mathrm{LN}(x^\prime_\ell))) = \sigma_{\mathrm{FFN}}^2 $. Then we have:

\begin{equation}
\sigma^2_{x_{\ell+1}} = \sigma_(x^\prime_\ell)^2 + \sigma_{\mathrm{FFN}}^2 +  \rho_2 \cdot \sigma_{x^\prime_\ell} \cdot  \sigma_{\mathrm{FFN}},
\end{equation}

where $\rho_2$ is the correlation factor. Thus, the relationship between $ \mathrm{Var}(x_{\ell+1}) $ and $ \mathrm{Var}(x_\ell) $ becomes:
\begin{equation}
\sigma^2_{x_{\ell+1}} =\sigma^2_{x_\ell} + \sigma_{\text{Attn}}^2 + \sigma_{\mathrm{FFN}}^2 + \rho_1 \cdot \sigma_{x_\ell} \cdot  \sigma_{\text{Attn}} + \rho_2 \cdot \sigma_{x^\prime_\ell} \cdot  \sigma_{\mathrm{FFN}}.
\label{variance_l_1}
\end{equation}

\subsubsection{Variance of the Attention}
The scaled dot-product attention mechanism is defined as:

\begin{equation}
\mathrm{Attn}(Q, K, V) = \mathrm{softmax}\left(\frac{QK^T}{\sqrt{d_k}}\right)V.
\end{equation}

The softmax function outputs a probability distribution over the keys. Let the softmax output be $A = \mathrm{softmax}\left(\frac{QK^T}{\sqrt{d_k}}\right)$, where $A$ is a matrix with each row summing to 1. The final attention output is obtained by multiplying the softmax output $A$ with the value matrix $V$:
\begin{equation}
   \mathrm{Attn}(Q, K, V) = AV.
\end{equation}

\begin{lemma}[\cite{ledoux2001concentration}]\label{lemma:softmax-variance}
Let $\{X_i\}_{i=1}^N$ be independent and identically distributed random variables with mean $m$ and variance $\sigma^2 < \infty$. Define the softmax weights $p_i = \frac{e^{X_i}}{\sum_{j=1}^N e^{X_j}}, \quad \text{and let} \quad p = (p_1, \ldots, p_N).$
Then, as $N \to \infty$, with high probability, the softmax vector $p$ concentrates around the uniform distribution on $N$ elements. In particular,
\begin{equation}
\lim_{n \rightarrow \infty}\mathbb{E}\left[\sum_{i=1}^N \left(p_i - \frac{1}{n}\right)^2\right] = 0,
\end{equation}
which implies that the softmax output becomes asymptotically indistinguishable, in expectation, from the uniform distribution.
\end{lemma}

According to the above lemma, to simplify the analysis, we make the following additional assumptions: The softmax output $A$ is approximately uniform, meaning each element of $A$ is roughly $1/n$, where $n$ is the number of keys/values. Given this assumption, the variance of the attention is:

\begin{equation}
\mathrm{Var}(\mathrm{Attn}(Q, K, V)) \sim \mathrm{Var}(AV) = \frac{1}{n} \sum_{i=1}^n d_{\mathrm{head}} \mathrm{Var}(V_i) = \frac{1}{n} \cdot n \sigma_V^2 \cdot d_{\mathrm{head}} = d_{\mathrm{head}} \sigma_V^2 = \sigma_W^2 d.
\end{equation}
where $W$ is the universal weight matrix defined as before.

\subsubsection{Variance of the Feed-Forward Network}

The feed-forward network (FFN) in transformers typically consists of two linear transformations with a ReLU activation in between. The FFN can be written as:

\begin{equation}
\mathrm{FFN}(x) = W_2 \cdot \mathrm{ReLU}(W_1 \cdot x + b_1) + b_2.
\end{equation}

where $W_1$ and $W_2$ are weight matrices, and $b_1$ and $b_2$ are bias vectors.

Using the result obtained by \citet{wang2024deepnet}, we get:

\begin{equation}
\sigma_{\mathrm{FFN}}^2 \sim \sigma_{W_1}^2 \cdot \sigma_{W_2}^2 = \sigma_W^4.
\end{equation}

In conclusion:

\begin{equation}
\begin{aligned}
\sigma^2_{x^\prime_\ell} &=\sigma_{x_\ell}^2 + \sigma_W^2 + \rho_2 \cdot \sigma_{x_\ell} \cdot  \sigma_W \\
&= \sigma_{x_\ell}^2 ( 1+ \frac{\sigma_W}{\sigma_{x_\ell}}+ \frac{\sigma_W^2}{\sigma^2_{x_\ell}})\\
& = \sigma_{x_\ell}^2 \Theta( 1 + \frac{1}{\sigma_{x_\ell}}).
\label{prime_or_not}
\end{aligned}
\end{equation}
For simplicity, we set the numerator part to 1. Substitute $
\sigma_{x^\prime_\ell} = \sigma_{x_\ell} \sqrt{1 + \frac{\sigma_W^2}{\sigma_{x_\ell}^2} + \rho_2 \cdot \frac{\sigma_W}{\sigma_{x_\ell}}}.
$ into Equation \eqref{variance_l_1} we get:
\begin{equation}
\begin{aligned}
\sigma^2_{x_{\ell+1}} &=\sigma^2_{x_\ell} +  \sigma_W^2 +  \sigma_W^4 d^2 + \rho_1 \cdot \sigma_{x_\ell} \cdot \sigma_W + \rho_2 \cdot \sigma_{x^\prime_\ell} \cdot  \sigma_W^2 d\\
& = \sigma^2_{x_\ell} + \sigma_W^2 + \sigma_W^4 d^2 + \rho_1 \cdot \sigma_{x_\ell} \cdot \sigma_W + \rho_2 \cdot \sigma_{x_\ell} \cdot \sigma_W^2 d+ \frac{\rho_2 \sigma_W^4 d^2}{2\sigma_{x_\ell}} + \frac{\rho_2^2 \sigma_W^3 d \sigma_{x_\ell}}{2} \\
& = \sigma_{x_\ell}^2 \Theta( 1 + \frac{1}{\sigma_{x_\ell}}).
\end{aligned}
\end{equation}

From the result we can generally infer that the variance accumulates layer by layer. The variance with regard to $\sigma_{x_1}$:
\begin{equation}
\sigma^2_{x_{\ell}} = \sigma_{x_1}^2 \Theta\Bigl(\prod_{k=1}^{\ell-1} \left( 1 + \frac{1}{\sigma_{x_k}} \right) \Bigr).
\end{equation}
We can also obtain a similar result for $\sigma^2_{x^\prime_{\ell}}$.

We observe that for any $ {\sigma^2_{x_k}} \geq 1 $, the sequence is increasing, meaning each term in the product is bounded. Consequently, the entire product is bounded above by:

\begin{equation}
{\sigma^2_{x_\ell}} \leq {\sigma^2_{x_1}} \prod_{k=1}^{\ell-1} \Bigl(1+\sqrt{\frac{1}{\sigma_{x_1}}} \Bigr)  = \sigma^2_{x_1}\bigl(1+\sqrt{\frac{1}{\sigma_{x_1}}} \Bigr) ^{\ell-1} = \exp{\Theta(L)}.
\label{variance_upper_bound}
\end{equation}

Taking the natural logarithm of both sides:

\begin{equation}
\begin{aligned}
\log({\sigma^2_{x_\ell}}) &= \log \left( \sigma_{x_1}^2 \prod_{k=1}^{\ell-1} \left( 1 + \sqrt{\frac{1}{{\sigma^2_{x_k}}}} \right) \right)
= \sum_{k=1}^{\ell-1} \log \left( 1 + \sqrt{\frac{1}{{\sigma^2_{x_k}}}} \right)+ \log(\sigma_{x_1}^2) \\
&\geq \sum_{k=1}^{\ell-1} \Bigl(\sqrt{\frac{1}{{\sigma^2_{x_k}}}} - \frac{1}{2} \left( \sqrt{\frac{1}{{\sigma^2_{x_k}}}} \right)^2 \Bigr)+ \log(\sigma_{x_1}^2).
\end{aligned}
\end{equation}

Exponentiating both sides to find the lower bound for  $\sigma^2_{x_\ell}$, we obtain:
$$
{\sigma^2_{x_\ell}} \geq \sigma_{x_1}^2 \exp \left( \sum_{k=1}^{\ell-1} \left( \sqrt{\frac{1}{{\sigma^2_{x_k}}}} - \frac{1}{2{\sigma^2_{x_k}}} \right) \right).
$$

This provides a tighter lower bound for ${\sigma^2_{x_\ell}} $ compared to the upper bound of Equation \eqref{variance_upper_bound}. Since we know the upper bound of variance grows exponentially, the lower bound must be sub-exponential. Therefore, for ${\sigma^2_{x_\ell}} = \ell $, we must have:

$$
\sigma^2_{x_\ell} \geq \sigma_{x_1}^2 \exp \left( \sum_{k=1}^{\ell-1} \left( \frac{1}{k} - \frac{1}{2k} \right) \right) = \Theta(\exp (\sqrt{L}))\geq \Theta(L).
$$
\end{proof}
Therefore, the increasing lower bound for ${\sigma^2_{x_\ell}}$ must grow faster than a linear function. So, the increase of variance is sub-exponential.
A large increase in such bound will lead to gradient spikes, which can connect to previous studies in \citet{huang2025stable,huang2025spam}.

\subsection{Proof of Theorem \ref{main_result}}
\label{appendix:proof2}
In this proof, we will divide the argument into two parts: first, the calculation of the Lemma \ref{upper_bound}, and second, the analysis of $\frac{\partial y_\ell}{\partial x_1}$.

\begin{lemma}
For an $L$-layered Pre-LN Transformer, $\frac{\partial y_L}{\partial x_1}$ using Equations \eqref{x_result} and \eqref{xprime_result} is given by:
\begin{equation}
\frac{\partial y_L}{\partial x_1} = \prod_{n=1}^{L-1} \left( \frac{\partial y_\ell}{\partial x^\prime_\ell} \cdot \frac{\partial x^\prime_\ell}{\partial x_\ell} \right).
\end{equation}

The upper bound for the norm of $\frac{\partial y_L}{\partial x_1}$ is:
\begin{equation}
\begin{aligned}
&\left\| \frac{\partial y_L}{\partial x_1} \right\|_2 \leq \prod_{l=1}^{L-1} \Big(\Big( 1 + \frac{\sigma^2}{\sigma_{x^\prime_\ell} (\sqrt{d} + \sqrt{d_\mathrm{FFN}})^2} \Big)\\
&\times \Big( 1 + 2dh\left( \sqrt{s} + 2 + \frac{1}{\sqrt{s}} \right) \frac{\sigma^2}{\sigma_{x_\ell} }  \Big(\sigma^2 d \sqrt{ d_{\mathrm{head}}} + \left( 1+ \sqrt{d_{\mathrm{head}}/d} \right) \Big) \Big).
\label{eq:upper_bound}
\end{aligned}
\end{equation}
\label{upper_bound}
\end{lemma}

Here, $h$ denotes the number of heads, $s$ is the sequence length, and $d$, $d_{\mathrm{FFN}}$, and $d_{\mathrm{head}}$ are the dimension of the embedding, FFN layer and multi-head attention layer, respectively. The standard deviation of \(W_Q\), \(W_K\), \(W_V\), and \(W_\mathrm{FFN}\) at layer $\ell$ is \(\sigma\) based on Assumption \ref{total_assumption}.

\subsubsection{Proof of Lemma \ref{upper_bound}}

\begin{proof}
Our derivation follows results in \cite{takase2023spike}, specifically Equation (7), which provides an upper bound on the norm of $\frac{\partial y_\ell}{\partial x_1}$ as:
\begin{equation}
\begin{aligned}
    \left\| \frac{\partial y_\ell}{\partial x_1} \right\|_2 = \left\| \prod_{l=1}^{L-1} \frac{\partial y_\ell}{\partial x^\prime_\ell} \frac{\partial x^\prime_\ell}{\partial x_\ell} \right\|_2.
    \end{aligned}
    \label{total_loss_iteration}
\end{equation}
Thus, we can estimate the upper bound of the gradient norm of $\frac{\partial y_\ell}{\partial x_1}$ by analyzing the spectral norms of the Jacobian matrices for the FFN layer and the self-attention layer, namely, 
\begin{equation}
\text{FFN:} \left\| \frac{\partial y_\ell}{\partial x^\prime_\ell} \right\|_2 \quad \text{Attention:} \left\| \frac{\partial x^\prime_\ell}{\partial x_\ell} \right\|_2.
\label{Attn_FFN}
\end{equation}

We now derive an upper bound of $\|\frac{\partial y_\ell}{\partial x^\prime_\ell}\|_2$ as follows:
\begin{equation}
    \left\| \frac{\partial y_\ell}{\partial x^\prime_\ell} \right\|_2 \leq 1 + \left\| \frac{\partial \mathrm{FFN}(\mathrm{LN}(x^\prime_\ell))}{\partial \mathrm{LN}(x^\prime_\ell)} \right\|_2 \left\| \frac{\partial \mathrm{LN}(x^\prime_\ell)}{\partial x^\prime_\ell} \right\|_2.
\end{equation}

Let $\sigma_{w1_\ell}$ and $\sigma_{w2_\ell}$ be the standard deviations of $W^1_\ell$ and $W^2_\ell$, respectively. From Assumption \ref{total_assumption}, the spectral norms of $W^1_\ell$ and $W^2_\ell$ are given by their standard deviations and dimensions \citep{vershynin2018}, so wo have:
$$\|W_1\|_2 \sim \sigma_1 \sqrt{d + \sqrt{d_\mathrm{FFN}}}.$$. 

For simplicity, we assume that $d$, and $d_\mathrm{FFN}$ are equal, thus,
\begin{equation}
    \left\| \frac{\partial \mathrm{FFN}(\mathrm{LN}(x^\prime_\ell))}{\partial \mathrm{LN}(x^\prime_\ell)} \right\|_2 = \|W^1_\ell W^2_\ell\|_2 \leq \sigma_1 \sigma_2 (\sqrt{d} + \sqrt{d_\mathrm{ffn}})^2.
\end{equation}

Finally, we have the following bound:
\begin{equation}
    \left\| \frac{\partial y_\ell}{\partial x^\prime_\ell} \right\|_2 \leq 1 + \frac{\sigma_{w1_\ell} \sigma_{w2_\ell}}{\sigma_{x^\prime_\ell} (\sqrt{d} + \sqrt{d_\mathrm{FFN}})^2} = 1 + \frac{\sigma^2_{\ell}}{\sigma_{x^\prime_\ell} (\sqrt{d} + \sqrt{d_\mathrm{FFN}})^2}. \label{pre_LN_FFN_result}
\end{equation}

Following a similar procedure for the FFN, we rewrite $\|\frac{\partial x'}{\partial x}\|_2$ in Equation \eqref{Attn_FFN} as:
\begin{equation}
\left\| \frac{\partial x'}{\partial x} \right\|_2 
\leq 1 + 
\left\| \frac{\partial \text{Attn}(\mathrm{LN}(x))}{\partial \mathrm{LN}(x)} \right\|_2 
\left\| \frac{\partial \mathrm{LN}(x)}{\partial x} \right\|_2.
\label{pre_LN_attn_result}
\end{equation}

Let $Z(\cdot) = \mathrm{concat}(\mathrm{head}_1(\cdot), \dots, \mathrm{head}_h(\cdot))$ and $J^Z$ denote the Jacobian of the $Z(\cdot)$. We can now express the spectral norm of the Jacobian matrix of attntion as:
\begin{equation}
\begin{aligned}
    \left\| \frac{\partial \text{Attn}(\mathrm{LN}(x_\ell))}{\partial \mathrm{LN}(x_\ell)} \right\|_2 = \left\| W^O_\ell Z(\mathrm{LN}(x_\ell)) \frac{\partial Z(\mathrm{LN}(x_\ell))}{\partial \mathrm{LN}(x_\ell)} \right\|_2 
    = \|W^O_\ell J^Z_\ell\|_2.
    \end{aligned}
\end{equation}

From \cite{vershynin2018}, we know that:
\begin{equation}
\begin{aligned}
 \|J^Z_\ell\|_2 \leq h \Big( \left( \sqrt{s} + 2 + \frac{1}{\sqrt{s}} \right) \sigma^3 \sqrt{d^3 d_{\mathrm{head}}} +\sigma^\ell_{x} \left( \sqrt{d}+ \sqrt{d_{\mathrm{head}}} \right) \Big).
 \end{aligned}
 \label{J_z}
\end{equation}
Here $h$ is the number of heads, $s$ is the sequence length, and the standard deviation of \(W_Q\), \(W_K\), and \(W_V\) is \(\sigma\).

By combining the inequalities \eqref{pre_LN_FFN_result}, \eqref{J_z} and \eqref{pre_LN_attn_result}, and assuming that all $\sigma$ values are the same for simplicity. we obtain: 

\begin{equation}
\begin{aligned}
&\left\| \frac{\partial y_L}{\partial x_1} \right\|_2 \leq \prod_{l=1}^{L-1} \Big(\Big( 1 + \frac{\sigma^2}{\sigma_{x^\prime_\ell} (\sqrt{d} + \sqrt{d_\mathrm{FFN}})^2} \Big)\\
&\times \Big( 1 + 2dh\left( \sqrt{s} + 2 + \frac{1}{\sqrt{s}} \right) \frac{\sigma^2}{\sigma_{x_\ell} }  \Big(\sigma^2 d \sqrt{ d_{\mathrm{head}}} + \left( 1+ \sqrt{d_{\mathrm{head}}/d} \right) \Big) \Big).
\end{aligned}
\end{equation}
\end{proof}

\subsubsection{Analysis of the Upper Bound}

As discussed in \cite{takase2023spike}, \(\sigma\) should be sufficiently small, and the standard deviation, \(\sigma_{x^\prime_\ell}\) or \(\sigma_{x_\ell}\) should satisfy the condition \(\sigma^2 \ll \sigma_{x^\prime_\ell}\) to maintain the lazy training scheme. Thus, we obtain the following bound for the product over $\ell$ from 1 to $L$:

To find the bound for $\left\| \frac{\partial y_{\ell}}{\partial x_1} \right\|_2$ with respect to $\ell$, we simplify the given inequality by approximating $\sigma_{x_\ell}$ and $\sigma_{x^\prime_\ell}$. Based on Equation \eqref{prime_or_not}, $\sigma_{x_\ell}$ is only one layer ahead of $\sigma_{x^\prime_\ell}$, and this layer does not significantly affect the overall performance of deep Transformer networks. Furthermore, based on Lemma \ref{convergence_of_variance}, we assume that $\sigma_{x^\prime_\ell} = \sigma_{x_\ell}$.

Equation \eqref{upper_bound} can be expressed in a traditional product form \cite{Whittaker_Watson_1996} for $\sigma_{x_\ell}$:

\begin{equation}
\left\| \frac{ \partial y_L}{\partial x_1} \right\|_2 \leq \prod_{l=1}^{L-1} \left( 1 + \frac{1}{\sigma_{x_\ell}} A + \frac{1}{\sigma_{x_\ell}^2} B \right),
\end{equation}

where

\begin{equation}
A = \frac{\sigma^2}{(\sqrt{d} + \sqrt{d_\mathrm{FFN}})^2} + 2dh \left( \sqrt{s} + 2 + \frac{1}{\sqrt{s}} \right) \sigma^2 \left( d \sqrt{d_{\mathrm{head}}} + 1 + \sqrt{d_{\mathrm{head}}/d} \right),
\label{A_form}
\end{equation}

and

\begin{equation}
B = 2dh \left( \sqrt{s} + 2 + \frac{1}{\sqrt{s}} \right) \sigma^4 d \sqrt{d_{\mathrm{head}}},
\label{B_form}
\end{equation}

where $A$ and $B$ are independent of $\sigma_{x_\ell}$, and under our assumption, are treated as constants. 

From classical infinite series analysis, it is known that as $\sigma_{x_\ell}$ grows at a faster rate, the upper bound of the product decreases. The proof is omitted here for brevity. For the upper bound on the convergence rate of $\sigma^2_{x_\ell}$, we assume $\sigma^2_{x_\ell} = \exp(\ell)$ without loss of generality. Under this condition, we can derive the following result:

Taking the natural logarithm of the product:

$$
\log \left( \prod_{k=1}^{L-1} \left(1 + \frac{A}{e^k} + \frac{B}{e^{2k}}\right) \right) = \sum_{k=1}^{L-1} \log \left(1 + \frac{A}{e^k} + \frac{B}{e^{2k}}\right).
$$

Using the Taylor series expansion for $\log(1 + x)$, and applying this to our sum, we get:
$$
\sum_{k=1}^{\infty} \log \left(1 + \frac{A}{e^k} + \frac{B}{e^{2k}}\right) = \sum_{k=1}^{\infty} \left( \frac{A}{e^k} + \frac{B}{e^{2k}} - \frac{1}{2} \left(\frac{A}{e^k} + \frac{B}{e^{2k}}\right)^2 + \frac{1}{3} \left(\frac{A}{e^k} + \frac{B}{e^{2k}}\right)^3 - \cdots \right).
$$
By evaluating the sums for each order of terms, we find that the result is a constant. Carrying this out for each term, we obtain:

$$
\log \left( \prod_{k=1}^{L-1} \left(1 + \frac{A}{e^k} + \frac{B}{e^{2k}}\right) \right) \sim \frac{A}{e-1} + \frac{B}{e^2 - 1} - \frac{1}{2} \left( \frac{A^2}{e^2 - 1} + 2 \frac{A \cdot B}{e^3 - 1} + \frac{B^2}{e^4 - 1} \right).
$$

Thus, the product is approximately:

\begin{equation}
\left\| \frac{\partial y_L}{\partial x_1} \right\|_2 \leq \exp \left( \frac{A}{e-1} + \frac{B}{e^2 - 1} - \frac{1}{2} \left( \frac{A^2}{e^2 - 1} + 2 \frac{A \cdot B}{e^3 - 1} + \frac{B^2}{e^4 - 1} \right) \right) = M,
\label{upper_constant}
\end{equation}
where $M$ is a constant.

For the lower bound on the convergence rate of $\sigma^2_{x_\ell}$, we assume $\sigma^2_{x_\ell} = \ell$ without loss of generality. Under this condition, we derive the following result. Taking the logarithm of the product, applying the Taylor series expansion for $\log(1 + x)$, and applying this to our sum:

$$
\sum_{k=1}^{\infty} \log \left(1 + \frac{A}{k} + \frac{B}{e^{k^2}}\right) = \sum_{k=1}^{\infty} \left( \frac{A}{k} + \frac{B}{e^{k^2}} - \frac{1}{2} \left(\frac{A}{k} + \frac{B}{e^{k^2}}\right)^2 + \frac{1}{3} \left(\frac{A}{k} + \frac{B}{e^{k^2}}\right)^3 - \cdots \right).
$$

For the first-order terms:

$$
\sum_{k=1}^{\infty} \left( \frac{A}{k} + \frac{B}{e^{k^2}} \right) = A \sum_{k=1}^{\infty} \frac{1}{k} + B \sum_{k=1}^{\infty} \frac{1}{e^{k^2}}.
$$

The series $\sum_{k=1}^{\infty} \frac{1}{k}$ is the harmonic series, which diverges. However, we approximate it using the Euler-Mascheroni constant $\gamma$ and the fact recognize that the harmonic series grows logarithmically:

$$
\sum_{k=1}^{\infty} \frac{1}{k} \sim \log n + \gamma \quad \text{(for large } n\text{)}.
$$

The other series such as $\sum_{k=1}^{\infty} \frac{1}{e^{k^2}}$ converge because $e^{k^2}$ grows very rapidly.

For higher-order terms, they converge to constant, involving the series $\sum_{k=1}^{\infty} \frac{1}{k^2}$ converges to $\frac{\pi^2}{6}$, so they contribute a constant. Exponentiating both sides, we get:

$$
\prod_{k=1}^{\infty} \left(1 + \frac{A}{k} + \frac{B}{e^{k^2}}\right) \sim \exp \left( A (\log n + \gamma) + const \right). 
$$

Thus, the growth rate of the upper bound for $\left\| \frac{\partial y_L}{\partial x_1} \right\|_2$ is:

\begin{equation}
    \left\| \frac{\partial y_L}{\partial x_1} \right\|_2 \leq \Theta(L).
    \label{delta_upper}
\end{equation}

\subsection{Proof of Lemma \ref{scaling_attention_variance}}
\label{appendix:proof3}
\begin{proof}

After scaling, the equation becomes:
\begin{equation}
\begin{aligned}
    y &= x_{\ell+1} = x^\prime_\ell + \mathrm{FFN}( \frac{1}{\sqrt{\ell}}\mathrm{LN}(x^\prime_\ell)),  \\
    x^\prime_\ell &= x_\ell + \mathrm{Attn}(\frac{1}{\sqrt{\ell}}\mathrm{LN}(x_\ell)). 
\end{aligned}
\end{equation}

Following the same analysis as before, we scale the Attention and FFN sub-layers, yielding:
\begin{equation}
\sigma_\mathrm{Attn}^2 = \frac{1}{n \ell} \cdot n \cdot \sigma_V^2 = \frac{1}{\ell} \sigma_V^2 = \frac{\sigma_W^2}{\ell},
\quad
\sigma_{\mathrm{FFN}}^2 \sim \frac{\sigma_{W_1}^2}{\ell} \cdot \frac{\sigma_{W_2}^2}{\ell} = \frac{\sigma_W^4}{\ell^2}.
\end{equation}

In conclusion:
\begin{equation}
\begin{aligned}
\sigma^2_{x^\prime_\ell} =\sigma_{x_\ell}^2 + \sigma_W^2 + \rho_2 \cdot \sigma_{x_\ell} \cdot  \frac{\sigma_W}{\sqrt{\ell}} = \sigma_{x_\ell}^2 \Theta( 1 + \frac{1}{\sqrt{\ell} \sigma_{x_\ell}}).
\label{prime_or_not_scaling}
\end{aligned}
\end{equation}
Similarly, we obtain:
\begin{equation}
\begin{aligned}
\sigma^2_{x_{\ell+1}} =\sigma_{x_\ell}^2 \Theta( 1 + \frac{1}{\sqrt{\ell} \sigma_{x_\ell}}).
\end{aligned}
\end{equation}

From the result we can generally infer that the variance accumulates layer by layer. The variance with regard to $\sigma_{x_1}$:
\begin{equation}
\begin{aligned}
\sigma^2_{x_{\ell}} =\sigma_{x_1}^2 \Theta\Big( \prod^{\ell-1}_{k=1}\Big(1 + \frac{1}{\sqrt{k} \sigma_{x_k}}\Big)\Big),
\end{aligned}
\end{equation}
We can also obtain a similar result for $\sigma^2_{x^\prime_{\ell}}$.

Taking the natural logarithm of both sides:

\begin{equation}
\begin{aligned}
\log({\sigma^2_{x_\ell}}) &= \log \left( \sigma_{x_1}^2 \prod_{k=1}^{\ell-1} \left( 1 + \sqrt{\frac{1}{{ k\sigma^2_{x_k}}}} \right) \right)
= \sum_{k=1}^{\ell-1} \log \left( 1 + \sqrt{\frac{1}{{k\sigma^2_{x_k}}}} \right)+ \log(\sigma_{x_1}^2) \\
&\geq \sum_{k=1}^{\ell-1} \Bigl(\sqrt{\frac{1}{{k\sigma^2_{x_k}}}} - \frac{1}{2} \left( \sqrt{\frac{1}{{k\sigma^2_{x_k}}}} \right)^2 \Bigr)+ \log(\sigma_{x_1}^2).
\end{aligned}
\end{equation}

To establish a lower bound for $ {\sigma^2_{x_\ell}} $, we exponentiate both sides. Setting  ${\sigma^2_{x_\ell}} = \ell $, we must have:

\begin{equation}
\sigma^2_{x_\ell} \geq \sigma_{x_1}^2 \exp \left( \sum_{k=1}^{\ell-1} \left( \frac{1}{k} - \frac{1}{2k} \right) \right) = \Theta(\exp (\log{L}))\geq \Theta(L).
\end{equation}
Therefore, the increasing lower bound ${\sigma^2_{x_\ell}}$ is greater than a linear function. 

Similarly, assuming ${\sigma^2_{x_\ell}} = \ell^{(2-\epsilon)} $, we have:
\begin{equation}
\begin{aligned}
\sigma^2_{x_\ell} &=  \sigma_{x_1}^2 \prod_{k=1}^{\ell-1} \left( 1 + \frac{1}{k^{2- \epsilon/2}} \right) \sim\exp\left( \sum_{k=1}^{\ell-1} \frac{1}{k^{2- \epsilon/2}} \right)\sim \exp\left( \frac{\ell^{\epsilon/2 - 1} - 1}{\epsilon/2 - 1} \right) \\
&\leq \Theta(\ell^{(2-\epsilon)})\leq \Theta(\ell^2).
\end{aligned}
\end{equation}
Here $\epsilon$ is a small constant with  $1/2 \leq \epsilon < 1$. Therefore, the increasing upper bound of ${\sigma^2_{x_\ell}}$ is slower than the $\ell^3$ function, leading to: 
$${\sigma^2_{x_\ell}} \leq \Theta(L^2)$$.

\end{proof}

\subsection{Proof of Theorem \ref{main_result_scaling}}
\label{appendix:proof4}
\begin{proof}
Similarly, after applying the scaling transformation, we derive an upper bound for $\|\frac{\partial y_\ell}{\partial x^\prime_\ell}\|_2$ as follows:
\begin{equation}
\begin{aligned}
    \left\| \frac{\partial y_\ell}{\partial x^\prime_\ell} \right\|_2 &\leq 1 + \left\| \frac{\partial \mathrm{FFN}(\mathrm{LN}(x^\prime_\ell))}{\partial \mathrm{LN}(x^\prime_\ell)} \right\|_2  \left\|  \frac{1}{\sqrt{\ell}} \right\|_2 \left\| \frac{\partial \mathrm{LN}(x^\prime_\ell)}{\partial x^\prime_\ell} \right\|_2 \\
    &= 1 + \frac{\sigma^2_{\ell}}{\ell \sigma_{x^\prime_\ell} (\sqrt{d} + \sqrt{d_\mathrm{FFN}})^2}.
     \end{aligned}
     \label{pre_LN_FFN_result_scaling}
\end{equation}

Similarly, rewriting Equation \eqref{Attn_FFN} after scaling, we have
\begin{equation}
\left\| \frac{\partial x'}{\partial x} \right\|_2 
\leq 1 + 
\left\| \frac{\partial \text{Attn}(\mathrm{LN}(x))}{\partial \mathrm{LN}(x)} \right\|_2 \left\|  \frac{1}{\sqrt{\ell}} \right\|_2
\left\| \frac{\partial \mathrm{LN}(x)}{\partial x} \right\|_2.
\label{pre_LN_attn_result_caling}
\end{equation}

By combining the bound \eqref{pre_LN_FFN_result_scaling}, and inequality \eqref{pre_LN_attn_result_caling}, and assuming all $\sigma$ are equal for simplicity, we obtain: 

\begin{equation}
\begin{aligned}
&\left\| \frac{\partial y_L}{\partial x_1} \right\|_2 \leq \prod_{l=1}^{L-1} \Big(\Big( 1 + \frac{\sigma^2}{\ell\sigma_{x^\prime_\ell} (\sqrt{d} + \sqrt{d_\mathrm{FFN}})^2} \Big)\\
&\times \Big( 1 + 2dh\left( \sqrt{s} + 2 + \frac{1}{\sqrt{s}} \right) \frac{\sigma^2}{\ell\sigma_{x_\ell} }  \Big(\sigma^2 d \sqrt{ d_{\mathrm{head}}} + \left( 1+ \sqrt{d_{\mathrm{head}}/d} \right) \Big) \Big).
\label{eq:upper_bound_scaling}
\end{aligned}
\end{equation}

Equation \eqref{eq:upper_bound_scaling} is a traditional product form \cite{Whittaker_Watson_1996} for $\sigma_{x_\ell}$. After scaling, it becomes:

\begin{equation}
\left\| \frac{ \partial y_L}{\partial x_1} \right\|_2 \leq \prod_{l=1}^{L-1} \left( 1 + \frac{1}{\ell\sigma_{x_\ell}} A + \frac{1}{\ell^2\sigma_{x_\ell}^2} B \right),
\end{equation}

where $A$ and $B$ retain their forms from Equation \eqref{A_form} and Equation \eqref{B_form} and are treated as constants. 

Regarding the upper bound on the convergence rate of $\sigma^2_{x_\ell}$, we assume $\sigma^2_{x_\ell} = \ell^{(2-\epsilon)}$ without loss of generality. For large $L$, the product can be approximated using the properties of infinite products:

\begin{equation}
\prod_{\ell=1}^{L-1} \left( 1 + \frac{A}{\ell^{2 - \epsilon/2}} + \frac{B}{\ell^{4 - \epsilon}} \right) \sim \exp\left( \sum_{\ell=1}^{L-1} \left( \frac{A}{\ell^{2 - \epsilon/2}} + \frac{B}{\ell^{4 - \epsilon}} \right) \right).
\end{equation}

Then, by evaluating the sum in the exponent, we obtain:

\begin{equation}
\prod_{\ell=1}^{L-1} \left( 1 + \frac{A}{\ell^{2 - \epsilon/2}} + \frac{B}{\ell^{4 - \epsilon}} \right) \sim \exp\left( A \cdot \frac{\ell^{\epsilon/2 - 1} - 1}{\epsilon/2 - 1} + B \cdot \frac{\ell^{\epsilon - 3} - 1}{\epsilon - 3} \right).
\end{equation}

Therefore, we establish the upper bound:
\begin{equation}
    \left\| \frac{ \partial y_L}{\partial x_1} \right\|_2 \leq \Theta \left( \exp\left( A \cdot \frac{\ell^{\epsilon/2 - 1} - 1}{\epsilon/2 - 1} + B \cdot \frac{\ell^{\epsilon - 3} - 1}{\epsilon - 3} \right)\right) = \omega(1),
\end{equation}
where $\omega(1)$ denotes a growth strictly greater than a constant as defined before.
\end{proof}

\subsection{Proof of theorem~\ref{thm:gradient-bound-lossspike}}\label{proof5}
\begin{proof}
We start with the Equation \eqref{xprime_result} and the chain‐rule:
\begin{equation}
\frac{\partial \mathcal{L}}{\partial W_1} = \frac{\partial \mathcal{L}}{\partial y_L}\frac{\partial y_L}{\partial x_1}\frac{\partial x_1}{\partial \mathrm{Attn}(\mathrm{LN}(x_1))}\frac{\partial \mathrm{Attn}(\mathrm{LN}(x_1))}{\partial \mathrm{Attn}(x_1)}\frac{\partial \mathrm{Attn}(x_1)}{\partial W_1}.
\end{equation}
where $\mathcal{L}$ is the loss function and $\frac{\partial \mathcal{L}}{\partial y_L}$ only relates to the composition of loss function.
So we only consider:

\begin{equation}
\frac{\partial y_L}{\partial W_1} 
= \frac{\partial y_L}{\partial x_1}\frac{\partial x_1}{\partial \mathrm{LN}(x_1)}\frac{\partial \mathrm{LN}(x_1)}{\partial \mathrm{Attn}(\mathrm{LN}(x_1))}\frac{\partial \mathrm{Attn}(\mathrm{LN}(x_1))}{\partial \mathrm{Attn}(x_1)}\frac{\partial \mathrm{Attn}(x_1)}{\partial W_1}.
\end{equation}

\begin{equation}
\left\|\frac{\partial y_L}{\partial W_1}\right\|_2 \le \left\|\frac{\partial y_L}{\partial x_1}\right\|_2 \cdot \left\|\frac{\partial x_1}{\partial \mathrm{LN}(x_1)}\right\|_2 \cdot \left\|\frac{\partial \mathrm{LN}(x_1)}{\partial \mathrm{Attn}(\mathrm{LN}(x_1))}\right\|_2 \cdot \left\|\frac{\partial \mathrm{Attn}(\mathrm{LN}(x_1))}{\partial \mathrm{Attn}(x_1)}\right\|_2 \cdot \left\|\frac{\partial \mathrm{Attn}(x_1)}{\partial W_1}\right\|_2.
\end{equation}
We know that:
\begin{equation}
   \left\|\frac{\partial x_1}{\partial\operatorname{LN}(x_1)}\right\|_2=\sigma_{x_1}.
\end{equation}

The paper \cite{vershynin2018} tells us that: 

\begin{equation}
   \left\|\frac{\partial \operatorname{Attn}(\operatorname{LN}(x_1))}{\partial \operatorname{LN}(x_1)}\right\|_2 
   = \| W^O_1 J^Z_1 \|_2.
\end{equation}

Now We want to calculate by writing the multi‐head attention (MHA) operation. Although various formulations exist, one common definition of MHA is as follows. For a given input x we compute

\begin{equation}
Q=xW^Q,\quad K=xW^K,\quad V=xW^V,
\end{equation}

and then for each head (we assume head index $i$ and $d_k$ the per‐head dimension) $\mathrm{head}_i = \operatorname{softmax}\Bigl(\frac{Q_iK_i^\top}{\sqrt{d_k}}\Bigr)V_i.$ The outputs of all h heads are then concatenated and projected with $W^O$: $\operatorname{Attn}(x)=\Bigl[\mathrm{head}_1,\ldots,\mathrm{head}_h\Bigr]W^O.$

$Q = xW_1,$ and then through $\mathrm{head} = \operatorname{softmax}\Bigl(\frac{QK^\top}{\sqrt{d_k}}\Bigr)V,$ so that (ignoring the outer projection $W^O$) we have

\begin{equation}
\operatorname{Attn}(x)\sim \operatorname{softmax}\Bigl(\frac{(xW_1)K^\top}{\sqrt{d_k}}\Bigr)V.
\end{equation}

So we have:
\begin{equation}
\left\|\frac{\partial\operatorname{Attn}(x)}{\partial W_1}\right\|_2\leq \frac{\|x\|_2}{\sqrt{d_k}},
\end{equation}

That is, writing it out explicitly,

\begin{equation}
\left\|\frac{\partial y_L}{\partial W_1}\right\|_2 \le \left\|\frac{\partial y_L}{\partial x_1}\right\|_2\sigma_{x_1}\|W^O_1J^Z_1\|_2\frac{\|x\|_2}{\sqrt{d}}.
\end{equation}
We know Equation \eqref{J_z}:
\begin{equation}
\begin{aligned}
 \|J^Z_\ell\|_2 \leq h \Big( \left( \sqrt{s} + 2 + \frac{1}{\sqrt{s}} \right) \sigma^3 \sqrt{d^3 d_{\mathrm{head}}} +\sigma^\ell_{x} \left( \sqrt{d}+ \sqrt{d_{\mathrm{head}}} \right) \Big).
 \end{aligned}
\end{equation}

Assume that $x \in \mathbb{R}^n$ is distributed as a multivariate normal with mean 0 and covariance $\sigma^2_1$. Then $\sum_{i=1}^n y_i^2$ follows a $\chi^2$ distribution with $n$ degrees of freedom. Thus,

\begin{equation}
\mathbb{E}\left[\|x\|_2\right] = \sqrt{\sigma^2_1}\mathbb{E}\left[\sqrt{\chi^2_n}\right] = \sqrt{\sigma^2_1}\sqrt{2}\frac{\Gamma\Bigl(\frac{n+1}{2}\Bigr)}{\Gamma\Bigl(\frac{n}{2}\Bigr)}.
\end{equation}

For large $n$, the chi-square distribution is concentrated around its mean ($n$) and one often approximates

\begin{equation}
\|x_1\|_2 \sim \sqrt{\sigma^2_1d}.
\end{equation}

\begin{equation}
\|W^O\|_2 \leq \sigma\Bigl(\sqrt{d}+\sqrt{hd_{\mathrm{head}}}\Bigr).
\end{equation}
Combine above together we have:

\begin{equation}
\left\|\frac{\partial y_L}{\partial W_1}\right\|_2 \le \left\|\frac{\partial y_L}{\partial x_1}\right\|_2(\sigma_{x_1})^2\|W^O_1\|_2h\left[\left(\sqrt{s}+2+\frac{1}{\sqrt{s}}\right)\sigma^3\sqrt{d^3d_{\mathrm{head}}}+\sigma_{x_1}\left(\sqrt{d}+\sqrt{d_{\mathrm{head}}}\right)\right].
\end{equation}

This is the desired upper bound for $\left\|\frac{\partial y_L}{\partial W_1}\right\|_2$.

Then we substitute the Equation \eqref{upper_bound} into the bound: and substituting the bounds for $\|W^O_1\|_2$ and $\left\|\frac{\partial y_L}{\partial x_1}\right\|_2$ into the original inequality yields

\begin{equation}
\begin{aligned}
\left\|\frac{\partial y_L}{\partial W_1}\right\|_2 &\le \left(\prod_{\ell=1}^{L-1} A_\ell\right) (\sigma_{x_1})^2  \sigma\left(\sqrt{d}+\sqrt{h d_{\mathrm{head}}}\right)  h \\
&\quad\quad \times \left[\left(\sqrt{s}+2+\frac{1}{\sqrt{s}}\right)\sigma^3\sqrt{d^3 d_{\mathrm{head}}} + \sigma_{x_1}\left(\sqrt{d}+\sqrt{d_{\mathrm{head}}}\right)\right].
\end{aligned}
\end{equation}

That is our final upper bound. For clarity, we summarize the answer:

\begin{equation}
\begin{aligned}
\left\|\frac{\partial y_L}{\partial W_1}\right\|_2 \le & \Big(\prod_{\ell=1}^{L-1} \left( 1 + \frac{\sigma^2}{\sigma_{x^\prime_\ell} (\sqrt{d}+\sqrt{d_{\mathrm{FFN}}})^2}\right)\Big( 1 + 2dh\left(\sqrt{s}+2+\frac{1}{\sqrt{s}}\right) \\
&\times \frac{\sigma^2}{\sigma_{x_\ell}} \left(\sigma^2 d\sqrt{d_{\mathrm{head}}} + \left(1+\sqrt{\frac{d_{\mathrm{head}}}{d}}\right)\right)\Big) \Big)\\
&\quad\times (\sigma_{x_1})^2  \sigma\left(\sqrt{d}+\sqrt{h d_{\mathrm{head}}}\right)  h \\
&\quad\times \left[\left(\sqrt{s}+2+\frac{1}{\sqrt{s}}\right)\sigma^3\sqrt{d^3 d_{\mathrm{head}}}+\sigma_{x_1}\left(\sqrt{d}+\sqrt{d_{\mathrm{head}}}\right)\right].
\end{aligned}
\end{equation}

So we find our that it is in the following form:
\begin{equation}
\left\| \frac{ \partial y_L}{\partial x_1} \right\|_2 \leq \prod_{l=1}^{L-1} \left( 1 + \frac{1}{\sigma_{x_\ell}} A^\prime + \frac{1}{\sigma_{x_\ell}^2} B^\prime \right),
\end{equation}
\end{proof}

\section{Variance Growth in Pre-LN Training}
\label{appendix:variance_analysis}

To analyze the impact of Pre-LN on variance propagation, we track the variance of layer outputs across different depths during training. 

\begin{figure*}[!ht]
\centering
\includegraphics[width=0.95\linewidth]{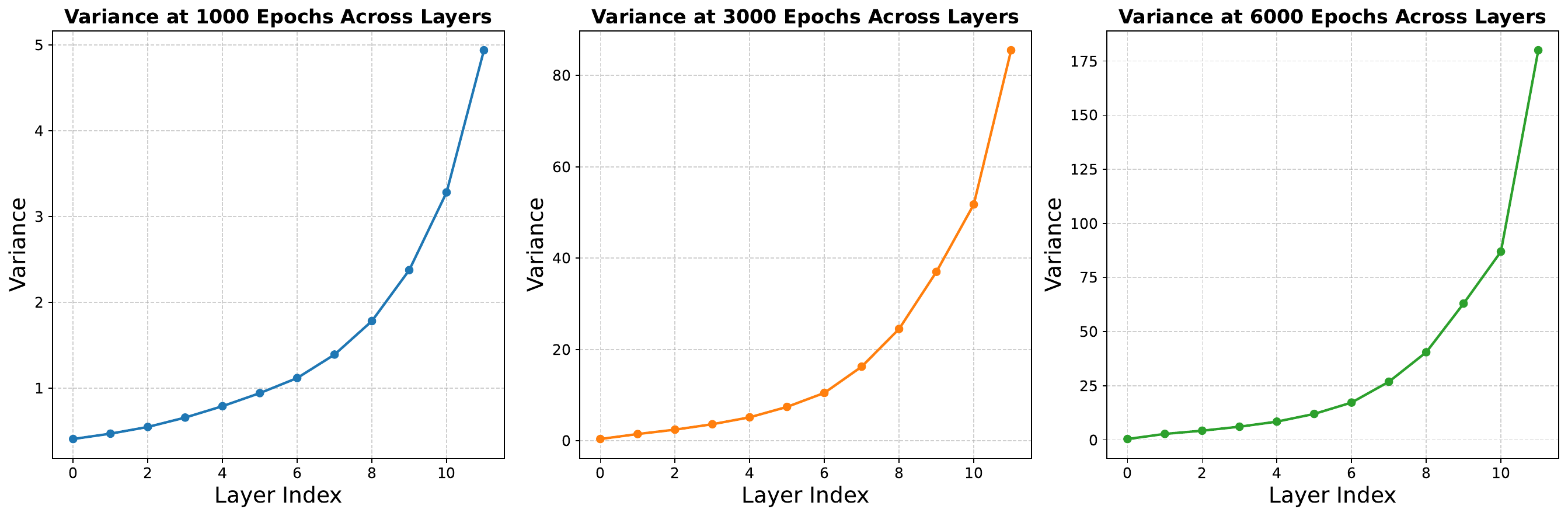}
\caption{
\textbf{Variance growth across layers in LLaMA-130M with Pre-LN.} 
Each subplot shows the variance at different training stages (1000, 3000, and 6000 epochs). 
In all cases, the variance follows an exponential growth pattern as depth increases, indicating that deeper layers experience uncontrolled variance amplification regardless of training progress.
}
\label{fig:variance_across}
\end{figure*}

Figure~\ref{fig:variance_across} illustrates the layer-wise variance in LLaMA-130M with Pre-LN at 1000, 3000, and 6000 epochs. Across all stages, variance remains low in shallow layers but grows exponentially in deeper layers, confirming that this issue persists throughout training rather than being a temporary effect.
This highlights the necessity of stabilization techniques like \ours to control variance and ensure effective deep-layer learning.

\section{Performance Drop of Layer Pruning in Vision–Language Models (Qwen 2.5-VL)}
\label{app:vlm}

\begin{figure}[!ht]
    \centering
    \includegraphics[width=\textwidth, height=0.28\textheight]{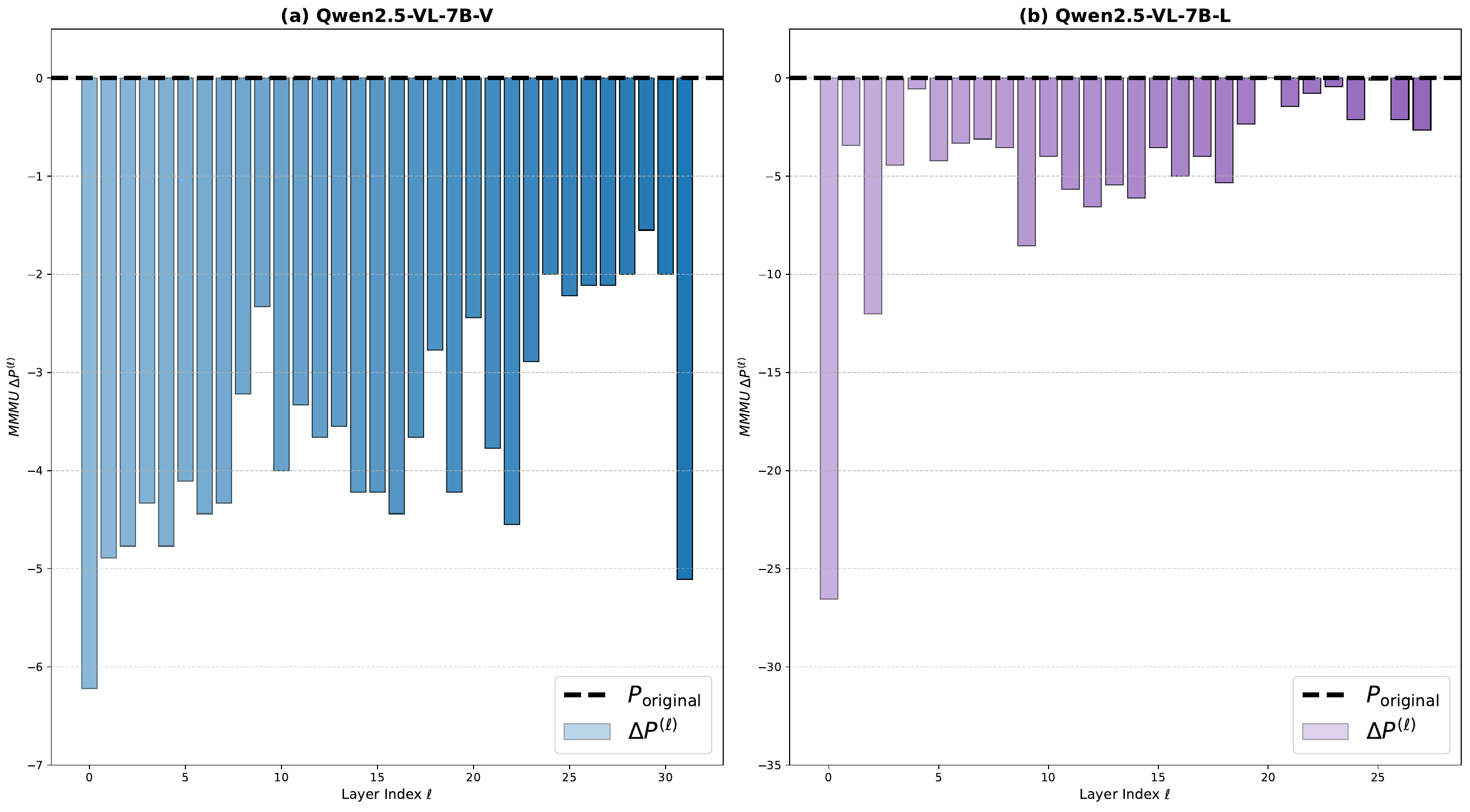}
    \caption{Performance drop of layer pruning on Qwen 2.5-VL-7B.  
    (a) Vision branch shows relatively uniform sensitivity across layers.  
    (b) Language branch exhibits a clear \textit{Curse of Depth}: deeper layers contribute much less than early ones.}
    \label{fig:qwen_vlm_prune}
\end{figure}

To examine whether the \textit{Curse of Depth} also manifests in vision–language models (VLMs), we perform layer–pruning experiments on \textbf{Qwen 2.5-VL-7B} ~\citep{bai2025qwen2}.  
For both its vision encoder and language decoder, we prune one transformer layer at a time and directly evaluate the pruned model on the MMMU benchmark ~\citep{yue2024mmmu}.  
Figure \ref{fig:qwen_vlm_prune} presents the resulting performance drops.

We observe that the \textbf{language branch} clearly suffers from the Curse of Depth, whereas the \textbf{vision branch} remains uniformly important.  
This suggests that the phenomenon is more pronounced in autoregressive language components of VLMs and may not directly transfer to vision encoders.  
A detailed modality–specific theoretical account is left to future work and community discussion.




\section{Limitations}

While this work offers a comprehensive analysis of the Curse of Depth in LLMs and proposes LayerNorm Scaling as an effective remedy, several limitations remain:

\paragraph{Scope of Architectures.} 
Our study primarily focuses on Transformer-based LLMs using Pre-LN. Although Pre-LN dominates modern architectures, our theoretical study does not cover models employing alternative normalization strategies (e.g., Post-LN only \citep{du2021glm}, normalization-free architectures \citep{zhu2025transformers}) or emerging paradigms such as mixture-of-experts or structured sparsity-based models.

\paragraph{Task Coverage.} 
Most empirical evaluations, including pruning and angular distance analyses, were conducted using general-purpose benchmarks like MMLU. While these tasks reflect broad model capabilities, domain-specific or long-context reasoning tasks may reveal different dynamics in deep layer contributions, which we leave for future work.

\paragraph{Fine-grained Representation Quality.} 
While LNS improves angular distance and performance sensitivity across layers, a deeper analysis of what types of information are represented or lost in deeper layers remains unexamined. For example, whether LNS helps preserve syntactic, semantic, or factual knowledge across depth is unclear.

\end{document}